# The Multi-Trip Autonomous Mobile Robot Scheduling Problem with Time Windows in a Stochastic Environment at Smart Hospitals

Lulu Cheng [1], Ning Zhao [1,*], Kan Wu [2] and Zhibin Chen [1]

[1] Faculty of Science, Kunming University of Science and Technology, Kunming 650500, China; chenglulu0206@163.com (L.C.); kan626@gmail.com (K.W.); chenzhibin311@126.com (Z.C.)
[2] Business Analytics Research Center, Chang Gung University, Taoyuan City 33302, Taiwan
* Correspondence: zhaoning@kust.edu.cn

**Abstract:** Autonomous mobile robots (AMRs) play a crucial role in transportation and service tasks at hospitals, contributing to enhanced efficiency and meeting medical demands. This paper investigates the optimization problem of scheduling strategies for AMRs at smart hospitals, where the service and travel times of AMRs are stochastic. A stochastic mixed-integer programming model is formulated to minimize the total cost of the hospital by reducing the number of AMRs and travel distance while satisfying constraints such as AMR battery state of charge, AMR capacity, and time windows for medical requests. To address this objective, some properties of the solutions with time window constraints are identified. The variable neighborhood search (VNS) algorithm is adjusted by incorporating the properties of the AMR scheduling problem to solve the model. Experimental results demonstrate that VNS generates high-quality solutions. Both enhanced efficiency and the meeting of medical demands are achieved through intelligently arranging the driving routes of AMRs for both charging and service requests, resulting in substantial cost reductions for hospitals and enhanced utilization of medical resources.

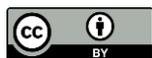

## 1. Introduction

Rapid advancements in artificial intelligence technology have facilitated the widespread integration of autonomous mobile robots (AMRs) into various domains, including industrial production [1] and medical rehabilitation. At smart hospitals, AMRs are employed to perform repetitive tasks, ensuring a safe distance between medical professionals and patients. This approach significantly mitigates the risk of personnel infection and enhances the overall safety of healthcare providers. Additionally, the utilization of AMRs in medical settings helps to alleviate the strain caused by staff shortages, thereby creating more value-added time for patient care [2].

This study is motivated by the specific requirements of Singapore Changi General Hospital, which intends to acquire a group of heterogeneous AMRs to improve medical services. The hospital has various routine medical requests, including tasks such as medication distribution, meal deliveries, and waste collection, which are intended to be handled by AMRs. Each type of medical request necessitates the use of a specific category of homogeneous AMRs. For instance, disinfection of wards requires different AMRs compared to those used for medication distribution. Consequently, the hospital needs to determine the optimal number of distinct AMR types to procure. It is essential to plan how these AMRs can deliver medical services effectively while adhering to multiple constraints [3].

The efficient and effective planning of service routes for AMRs is crucial for their operation. Scheduling optimization of AMRs is a type of research based on resource allocation that has gradually emerged in recent years. Previous studies have investigated various aspects of robot route planning. Dang et al. [4] examined the sequencing problem of a single robot delivering parts to production lines, minimizing total travel time while adhering to time windows. Building upon this work, Dang et al. [5] increased the capacity constraints during the service process. Booth et al. [6] considered the physical constraint of battery levels and incorporated charging during driving in a single robot task planning problem. Jun et al. [7] focused on scheduling and routing for AMR systems, aiming to minimize transportation request tardiness using a novel local search algorithm.

All of the above literature assumes one robot performs only one route/trip. However, in practical applications, small vans are commonly used for package delivery due to their high mobility. In addition, drones and AMRs have the potential for delivering essential goods without direct human contact, but their limited capacity may require multiple trips [8]. To improve service efficiency, multiple robots can be arranged to make multiple trips. In recent years, there has been significant growth in research publications regarding robot scheduling problems that involve multiple trips. Liu et al. [9,10] address a mobile robot routing problem with multiple trips and time windows, emphasizing strict adherence to time windows at each stop. They proposed a two-index mixed integer programming model to solve this problem. Yao et al. [11] investigated the scheduling of multiple mobile robots for multiple trips, aiming to minimize transportation and waiting costs. They employed a hybrid approach combining an improved genetic algorithm and tabu search algorithm to find optimal routes for the robots. Notably, [9–11] did not consider electrical constraints in the context of robot scheduling problems. Moreover, these studies assume a deterministic environment, while the stochastic nature of driving and service times can significantly impact performance, as discussed in [12,13]. This paper focuses on addressing these specific issues.

This paper examines the scheduling strategy for autonomous mobile robots (AMRs), using drug delivery as a case study. It addresses a multi-trip route scheduling problem specific to a particular type of AMR in a stochastic environment characterized by stochastic travel and service times. The study aims to improve service efficiency and minimize the number of required AMRs by considering the battery level, time window, and load capacity during the service process. The primary objectives of this problem are to minimize operational costs while fulfilling all service requests. To achieve cost reduction and improve AMR efficiency, a stochastic mixed-integer programming model is developed to minimize hospital investment costs, encompassing both the fixed and dynamic expenses associated with the AMRs. Although initially motivated by medical services, this problem has wide-ranging applications in other domains, such as city logistics and drone operations within large-scale transportation networks.

Given the stochastic nature of the scheduling problem for AMRs, this issue is addressed within the framework of stochastic programming. There are two commonly employed approaches in stochastic programming: the expected value model and the chance-constrained model [14]. For variables with less randomness, the mean values can replace the random variables in the objective function and constraints, allowing for the use of deterministic methods to solve the problem. Conversely, when dealing with highly random variables, the chance-constrained model is typically adopted, although it introduces some additional complexity due to the variable distributions. The chance-constrained model, proposed by Charnes and Cooper [14], is a powerful tool for modeling stochastic decision systems. It requires that the probability of the constraint condition is no less than a certain confidence level [15]. Ge et al. [16] explored the electric vehicle routing problem with stochastic demands by constructing a probability constraints model. In our study, given the stochastic travel time and service time, violations of request time windows are inevitable. Consequently, we develop a chance-constrained mixed-integer programming model in which the probability of an AMR arriving within the designated time window should be greater than a given threshold [17].

The AMR routing scheduling problem is an NP-hard combinatorial optimization problem [4]. Exact algorithms are only suitable for small- to medium-scale data instances of the routing problem. To tackle complex routing problems, research has primarily focused on developing heuristic algorithms that can generate high-quality solutions within reasonable computational times. The most common heuristic algorithms are the genetic algorithm [18], variable neighborhood search algorithm [19], ant colony algorithm [20], and tabu search algorithm [21,22]. Among these approaches, the variable neighborhood search heuristic (VNS) has emerged as an effective and competitive method for solving large-scale routing problems [19,23–25]. Initially proposed by Mladenovi and P. Hansen [26], VNS offers a framework for constructing heuristics by systematically exploring different neighborhoods while employing local search techniques to reach local optima. The VNS algorithm effectively mitigates the risk of getting trapped in local optima and rapidly identifies optimal solutions. Numerous variants have been derived from this fundamental algorithmic framework and applied to various combinatorial optimization problems such as vehicle routing [27,28], electric vehicle routing with stochastic demands [29], traveling salesman problems [30,31], and scheduling problems [32–34].

The VNS algorithm is particularly advantageous for handling large problem instances where local search procedures are time-consuming, and it can also serve as an effective means of generating initial solutions before decomposition. The algorithm operates by systematically exploring the solution space, using neighborhood operators to drive the search process toward promising regions. By iteratively applying these operators and conducting local searches, VNS aims to achieve better solutions than traditional optimization techniques while effectively navigating the complex landscape of the problem at hand. However, it is important to note that these operators may generate solutions that are not always feasible. In this study, we improve the VNS algorithm by incorporating time window constraints into three neighborhood solution generation operators. Moreover, we introduce two repair operators, namely depot insertion and charging insertion, which can transform any infeasible solution into a feasible one. Finally, we propose an AMR decrease operator to reduce the number of AMRs in the current set of feasible solutions.

The main contributions of this paper are as follows:

(1) A stochastic mixed-integer programming model is established to analyze the medical AMR routing problem with stochastic travel time and service time.
(2) To increase system utilization and decrease the number of AMRs required, each AMR can run multiple routes every day.
(3) The chance constraint of the violated time window is considered to guarantee a high service level.
(4) Based on the observed AMR service process, we introduce two repair operators: the depot insertion operator and the charging insertion operator. Additionally, an AMR decrease operator is implemented. These operators are incorporated into the VNS algorithm to address the stochastic AMR routing problem.

The remainder of this paper is structured as follows. We describe the problem in Section 2 and formulate the optimization model in Section 3. The computational method is presented in Section 4. Section 5 provides numerical examples and discusses the results. Section 6 concludes the paper.

## 2. Problem Description

The motivation for this study arises from the hospital's intention to acquire a fleet of AMRs to assist medical personnel in providing certain medical services. There are some daily medical requests to be completed by some homogeneous AMRs. The primary task of the AMRs involves transporting medicines from depots (e.g., pharmacies) to different demand points (e.g., wards) and returning to the depots upon completing the service. However, considering the diverse demands associated with each request, the service time for the AMRs is uncertain. Furthermore, during the driving process, multiple unpredictable factors influence travel time, such as obstacle avoidance or waiting for the elevator. As a result, both service time and travel time are treated as random variables. Each medical request is assigned a specific time window within which it must be completed, and the AMRs are responsible for delivering a specified quantity of demands for each request. It is important to note that each request is fulfilled by one and only one AMR. Assuming that the capacity of each AMR is denoted by $Q$, the AMRs consume electricity during their operations. To improve the efficiency of AMRs and minimize the hospital's daily costs, it becomes imperative to plan the optimal number of participating AMRs and devise the most efficient AMR service routes.

The following characteristics of the AMR service process should be taken into consideration:

(1) Stochastic travel time: due to unforeseeable obstacles, the travel time between any two requests is subject to randomness. Additionally, the AMR waiting time for elevators is also stochastic.
(2) Stochastic service time: the duration of AMR service varies depending on the service object or environment.
(3) Time window: medical requests are associated with specific time windows. Failure to deliver medicine to the patient within the designated timeframe can result in severe consequences.
(4) Battery constraint: low battery levels require immediate recharging at the charging station.
(5) Capacity constraint: the AMR's load capacity sets an upper bound on the items it can carry.

To facilitate the analysis of the routing problem of the medical AMRs, a complete directed graph $G=(V,A)$ is defined, where $V = D_1 \cup R \cup C \cup D_2$, $D_1$ is the set of starting points of the AMRs, $D_2$ is the set of endpoints, $C$ is the set of charging stations, and $R=\{1,2,3,...,n\}$ is the set of medical requests issued by the hospital. Let $A=\{(i,j)|i,j \in V, i \neq j\}$ be the set of arcs. If an AMR needs to charge during operation, it selects the nearest charging point to complete the charge. Without loss of generality, let $D_1 = D_2 = \{d\}$. Suppose that the time window of request $i$ is $[e_i, h_i]$. If an AMR arrives at the request point $i$ earlier than the lower bound of its time window $e_i$, it waits until $e_i$. To ensure service quality, the chance constraint is adopted. Let the probability that AMRs serve requests in their specific time window be greater than a given threshold $1-\varepsilon$. The service time of the request $i$ is a random variable $S_i$. The AMR needs to transport the corresponding demand $q_i$ to the request point $i$, $i \in R$. The distance between any two medical requests is $d_{ij}$, and their floor difference is $|f_{ij}|$, $(i,j) \in A$. In the stochastic environment there are various uncertainties, such as avoiding obstacles or taking elevators during the travel of AMRs. The travel time $T_{ij}$ between request $i$ and request $j$ is a random variable.

Figure 1 illustrates a simple example. It involves three AMRs, twelve medical requests, a charging point, and a depot (pharmacy). All AMRs depart from the depot and eventually return to the depot. The two elements in vector $(l_i\ level, q_i\ kg)$ represent the floor and the demand of the request point $i$, respectively.

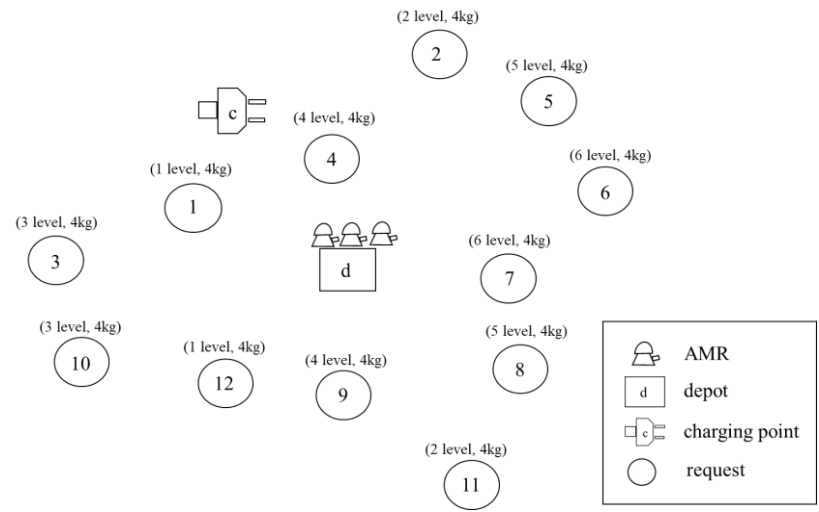

**Figure 1.** An illustrated example.

The parameters and variables are summarized in Table 1.

**Table 1.** Nomenclature list.

| Parameter | Description |
| --- | --- |
| $D_1 = \{d\}$ | the set of starting points of AMRs |
| $D_2 = \{d\}$ | the set of ending points of AMRs |
| $R$ | the set of request points |
| $C$ | the set of charging stations |
| $V_1 = D_1 \cup R \cup C$ | |
| $V_2 = D_2 \cup R \cup C$ | |
| $V = D_1 \cup R \cup C \cup D_2$ | |
| $m$ | number of AMRs participating in medical request services |
| $q_i$ | demand of request $i$ |
| $[e_i, h_i]$ | the time window of request $i$ |
| $Q$ | capacity of AMRs |
| $S_i$ | service time of request $i$ |
| $d_{ij}$ | the distance between any two medical requests $i$ and $j$ |
| $|f_{ij}|$ | their floor difference between any two medical requests $i$ and $j$ |
| $T_{ij}$ | the travel time from request $i$ to request $j$ |
| $X_{pi}^k$ | the time when the $k$th AMR arrives at the request $i$ on the $p$th route |
| $Y_{pi}^k$ | the time when the $k$th AMR starts service request $i$ on the $p$th route |
| $\alpha$ | the lower bound of the AMR battery level |
| $\beta$ | the upper bound of the AMR battery level |
| $r$ | electricity consumption rate, i.e., the AMR consumes $r$ percent of its fully charged battery level if it runs a unit time |
| $v_q$ | the charging rate of AMRs, i.e., the AMR increases $v_q$ percent of its fully charged battery level if it is charged for a second |
| $\gamma_{pi}^k$ | the remaining battery level of the $k$th AMR when it arrives at request (or depot) $i$ on its $p$th route |
| $u_{pi}^k$ | the remaining capacity of the $k$th AMR when it arrives at request $i$ on its $p$th route |
| $h_0$ | any infinite large number |
| $\xi_1$ | the daily fixed cost of each AMR |
| $\xi_2$ | the unit travel cost of the AMR |
| $\xi_3$ | the penalty cost for violating the time window |
| $\varepsilon$ | the probability that the time window is violated |
| $x_{pij}^k$ | binary decision variable, i.e., $x_{pij}^k = \begin{cases} 1, & \text{the } k\text{th AMR serves requests } i \text{ and } j \text{ successively on its } p\text{th route;} \\ 0, & \text{otherwise.} \end{cases}$ |

## 3. Model Formulation

From the perspective of hospital resource allocation, it is desirable to minimize the number of AMRs required to fulfill all medical requests. This necessitates enhancing the utilization of AMRs and devising optimal service routes. In this section, we formulate a mathematical programming model aimed at minimizing the economic cost incurred by the hospital.

Assume that the number of AMRs participating in medical services is $m$, where $m$ is a decision variable. Each AMR may drive multiple routes every day. Let the set of traveling routes of the $k$th AMR be $L_k = \{1, 2, \cdots\}$, $k = 1, \ldots, m$. Assume that $T_{ij}$ and $S_i$ follow normal distributions. The $k$th AMR arrives at request $i$ on its $p$th route at $X_{pi}^k$, and it starts to serve request $i$ at $Y_{pi}^k$. Moreover, $Y_{pi}^k = \max\{X_{pi}^k, e_i\}$ since the AMR needs to wait until $e_i$ if $X_{pi}^k < e_i$. If the $k$th AMR serves requests $i$ and $j$ successively on its $p$th route, the arrival time $X_{pj}^k$ is

$$X_{pj}^k = Y_{pi}^k + S_i + T_{ij}, \forall i, j \in V, k = 1, 2, \ldots, m, p \in L_k \tag{1}$$

Let the full battery capacity be 1, and the electricity consumption rate is a percentage of the battery capacity. The remaining battery level of the AMRs is denoted by a percentage of the fully charged battery level. Suppose that when the $k$th AMR reaches request $i$ on the $p$th route ($i \in R$), the remaining battery level is $\gamma_{pi}^k$, and the remaining capacity is $u_{pi}^k$. The electricity consumption rate per unit time of the AMRs is $r$. The AMR charging rate is $v_q$, which means that the AMR increases $v_q$ percent of its fully charged battery if it is charged for a second. The AMR is allowed to perform partial charging. When the battery level is lower than $\alpha$ the AMR needs to charge, and it can start to work when the battery level is greater than or equal to $\beta$, where $0 \leq \alpha < \beta \leq 1$.

For the above scheduling problem of medical AMRs, the objective is to minimize the operation cost of the hospital, including the fixed cost and dynamic cost of AMRs, where the fixed cost is the cost of AMRs and the dynamic cost is the energy consumption cost of AMRs during operation. The daily fixed cost of each AMR is $\xi_1$ and the unit travel cost of the AMR is $\xi_2$. For this problem, the following mixed-integer programming model is established. The constraints include the battery and capacity of the AMR, the medical request time window, etc.

Objective function:

$$\min f(m,x) = \xi_1 m + \xi_2 \sum_{k=1}^{m} \sum_{p \in L_k} \sum_{\substack{i \in V_1 \\ j \in V_2 \\ i \neq j}} d_{ij} x_{pij}^k \quad (2)$$

subject to:

$$\sum_{k=1}^{m} \sum_{p=1}^{l_{max}^k} \sum_{j \in V_2, i \neq j} x_{pij}^k = 1, \forall i \in R \quad (3)$$

$$\sum_{j \in V_2, j \neq i} x_{pij}^k = \sum_{j \in V_1, j \neq i} x_{pji}^k, \forall i \in V, k=1,2,...,m, p \in L_k \quad (4)$$

$$q_j \leq u_{pj}^k \leq u_{pi}^k - x_{pij}^k \cdot q_i + Q(1-x_{pij}^k), \forall i \in V_1, j \in V_2, i \neq j, k=1,2,...,m, p \in L_k \quad (5)$$

$$Y_{pi}^k + S_i + T_{ij} \cdot x_{pij}^k - h_0 \cdot (1-x_{pij}^k) \leq Y_{pj}^k, \forall i \in D_1 \cup R, j \in V_2, i \neq j, k=1,2,...,m, p \in L_k \quad (6)$$

$$Y_{pi}^k + \frac{\beta - \gamma_{pi}^k}{v_q} + T_{ij} - h_0 \cdot (1-x_{pij}^k) \leq Y_{pj}^k, \forall i \in C, j \in D_2 \cup R, k=1,2,...,m, p \in L_k \quad (7)$$

$$Y_{pi}^k = Y_{p+1,i}^k, \forall i \in D_1 \cap D_2, k=1,2,...,m, p \in L_k \quad (8)$$

$$Y_{pi}^k \geq e_i, \forall i \in R, k=1,2,...,m, p \in L_k \quad (9)$$

$$P\{X_{pi}^k \leq h_i\} \geq 1-\varepsilon, \forall i \in V, k=1,2,...,m, p \in L_k \quad (10)$$

$$\alpha \leq \gamma_{pj}^k \leq \gamma_{pi}^k - (r \cdot d_{ij}) \cdot x_{pij}^k + (1-x_{pij}^k), \forall i \in V_1, j \in V_2, i \neq j, k=1,2,...m, p \in L_k \quad (11)$$

$$\gamma_{pi}^k = \gamma_{p+1,i}^k, \forall i \in D_1 \cap D_2, k=1,2,...,m, p \in L_k \quad (12)$$

$$x_{pij}^k \in \{0,1\}, \forall i \in V_1, j \in V_2, i \neq j, k=1,2,...,m, p \in L_k \quad (13)$$

$$m \in N^+ \quad (14)$$

The objective function (2) is formed by three parts: the daily fixed cost, transportation cost, and the penalty cost of violating the time window. Equation (3) ensures that each request is served in one and only one route. Equation (4) indicates that the number of input arcs at each point is equal to the number of output arcs. The inequality (5) represents the capacity constraint of the $k$th AMR on its $p$th route. The inequality (6) is the time constraint of the *k*th AMR reaching request *j* from request *i* on its *p*th route. The inequality (7) is the time constraint for the *k*th AMR leaving charging station *i* and reaching request *j* on its *p*th route. Equation (8) means that the time when the *k*th AMR arrives at the ending point of the *p*th route is equal to the time when the *k*th AMR departs from the starting point of its $(p+1)$th route. The inequality (9) ensures that the AMR serves request *i* later than or equal to the lower bound of the time window. The inequality (10) represents the chance constraint which ensures that request *i* is served in its time window with a confidence level $1-\varepsilon$. The constraints (11) and (12) are about the battery level of AMRs. The formula (13) denotes a binary decision variable $x_{pij}^k$. The constraint (14) indicates that the variable *m* belongs to the set of positive integers.

The traditional method to solve the chance-constrained programming problem is by converting the chance constraints into deterministic equations. We adopt the approach of Ehmke et al. [35] to approximate the arrival time using a normal distribution, i.e., $X_{pi}^k \sim N(\mu(X_{pi}^k), \sigma^2(X_{pi}^k))$. Let $x_{pij}^k = 1$ ($i,j \in V$), then $X_{pj}^k = Y_{pi}^k + S_i + T_{ij}$, where the starting service time $Y_{pi}^k = \max\{X_{pi}^k, e_i\}$ follows a left-truncated non-normal distribution. From Nadarajah and Kotz [36], we have

$$\mu(Y_{pi}^k) = \mu(X_{pi}^k)\Phi\left((\mu(X_{pi}^k)-e_i)/\sigma(X_{pi}^k)\right) + e_i\Phi\left((e_i-\mu(X_{pi}^k))/\sigma(X_{pi}^k)\right) + \sigma(X_{pi}^k)\phi\left((\mu(X_{pi}^k)-e_i)/\sigma(X_{pi}^k)\right), \quad (15)$$

$$\sigma^2(Y_{pi}^k) = \left(\mu^2(X_{pi}^k)+\sigma^2(X_{pi}^k)\right)\cdot\Phi\left(\frac{\mu(X_{pi}^k)-e_i}{\sigma(X_{pi}^k)}\right) + \Phi\left(\frac{e_i-\mu(X_{pi}^k)}{\sigma(X_{pi}^k)}\right)\cdot e_i^2 - \left(\mu(Y_{pi}^k)\right)^2 + \left(\mu(X_{pi}^k)+e_i\right)\sigma(X_{pi}^k)\cdot\phi\left(\frac{\mu(X_{pi}^k)-e_i}{\sigma(X_{pi}^k)}\right), \quad (16)$$

where $\Phi(\cdot)$ is the cumulative distribution function of the standard normal distribution and $\phi(\cdot)$ is the probability density function of the standard normal distribution.

Since the travel time $T_{ij}$ and the service time $S_i$ are independent random variables, we have

$$\mu(X_{p,j}^k) = \mu(Y_{pi}^k) + \mu(S_i) + \mu(T_{ij}) \quad (17)$$

and

$$\sigma^2(X_{p,j}^k) = \sigma^2(Y_{pi}^k) + \sigma^2(S_i) + \sigma^2(T_{ij}). \quad (18)$$

Then constraint (10) $P\{X_{p,j}^k \leq h_j\} \geq 1-\varepsilon$ can be converted to

$$\mu(Y_{pi}^k) + \mu(S_i) + \mu(T_{ij}) + z_\varepsilon\sqrt{\sigma^2(Y_{pi}^k) + \sigma^2(S_i) + \sigma^2(T_{ij})} \leq h_j, \quad (19)$$

where $z_\varepsilon$ is the $\varepsilon$ quantile of the standard normal distribution.

## 4. Reduced Variable Neighborhood Search

The stochastic model established in Section 3 is known to be NP-hard and could be solved by heuristic algorithms [37]. In this section, we present the proposed algorithm variable neighborhood search (VNS) and subsequently introduce its four main components: initial solution, neighborhood change, shaking procedure, and feasible operation.

It is important to note that the feasible operation (depot insertion and charging insertion) guarantees that the generated solutions always satisfy the constraints of time windows, capacity, and battery. Additionally, the AMR decrease operator aims to minimize the number of AMRs in the current feasible neighborhood solutions.

The framework of the VNS is presented in Section 4.1. The initial solution of the VNS algorithm is introduced in Section 4.2. Section 4.3 presents the local search and shaking procedure.

*4.1. The Algorithm of the VNS*

Algorithm 1 clearly illustrates the overall framework of our proposed algorithm VNS, where $N$ represents the maximum number of iterations. First, the greedy algorithm is applied to obtain an initial solution $x$, which is then transformed into a feasible solution using the feasible operation and considered as the current best solution $x^*$ (lines 1–3). After that, VNS enters an evolutionary process. In each iteration (lines 5–11), $x$ undergoes a local search to generate a neighboring solution $x_l$. A shake method is utilized to move $x_l$ out of the local optimum and obtain a new neighboring solution $x_s$, resulting in an improved solution $x^*$. Finally, the output is the final solution $x^*$ (line 13). The time complexity of the algorithm is $O(N \cdot |R|^2)$, where $|R|$ represents the number of requests and $N$ is the maximum number of iterations.

---

**Algorithm 1.** The VNS Algorithm

**Input: parameters** $n \leftarrow 1$, $N$

**Output: the final solution** $x^*$

1. $x \leftarrow$ **Generate initial solution;**//see Section 4.2
2. $x \leftarrow$ **Feasible operation (** $x$ **);**//Algorithm 4
3. $x^* \leftarrow x$;
4. **While** $n \leq N$ **do**
5.     $x_l \leftarrow$ **Local search (** $x$ **);**//Algorithm 2
6.     $x_l \leftarrow$ **Feasible operation (** $x_l$ **);**//Algorithm 4
7.     $x_s \leftarrow$ **Shaking (** $x_l$ **);**//Algorithm 3
8.     **If** $f(x') < f(x^*)$ //select the best solution from $x'$ and $x^*$
9.         $x^* \leftarrow x'$
10.        $x \leftarrow x'$ //update the current solution
11.    **End if**
12. **End while**
13. **Return** $x^*$

---

*4.2. An Initial Solution*

The initial solution of the VNS is constructed by a greedy algorithm. Without loss of generality, it is assumed that each AMR can complete at least one request without violating the constraints of load capacity, battery level, and time windows. The initial solution is constructed in the following way.

The starting point for AMRs is the depot. In every iteration, requests near the current node are added to the current route at the most suitable insertion position. If a route reaches its capacity limit and cannot accommodate more requests, a new route is created.

When the remaining battery level goes below a specified threshold $\alpha$, a charging station is placed on the current route. This process continues until all customers have been visited. More details about depot insertion and charging station placement can be found in Section 4.3.

*4.3. Local Search and Shake Procedure*

The local search procedure is the core component of VNS, as shown in lines 4–11 of Algorithm 2. The neighborhood of a solution $x$ is denoted by $N_k(x)$, $k=1,\cdots,3$.

---

**Algorithm 2.** Local Search

**Input: the initialized solution** $x$, $k$ **neighborhoods** $N_k(x)$, $k=1,\cdots,3$.

**Output: a solution** $x_l$

1. $k \leftarrow 1$
2. $x' \leftarrow N_k(x)$
3. **Repeat**
4.     **If** $f(x') < f(x)$ **then**
5.         $x \leftarrow x'$ //update the current best solution
6.         $k \leftarrow 1$ //initial neighborhood
7.         $x' \leftarrow N_k(x)$ //update the current solution
8.     **Else**
9.         $k \leftarrow k+1$ //next neighborhood
10.        $x' \leftarrow N_k(x)$ //update the current solution
11.     **End if**
12. $x_l \leftarrow x$ //output the best solution
13. **Returns** $x_l$

---

It is commonly observed that the operators used for generating neighborhood solutions exhibit a stochastic nature, and the generated neighborhoods may include infeasible solutions or tend towards suboptimal outcomes due to constraint violations. In order to minimize the generation of infeasible neighboring solutions, particularly those violating time window constraints, we propose Proposition 1 and Proposition 2 for this model.

**Proposition 1.** *Let the time windows of two requests $i_1$ and $i_2$ be $\left[e_{i_1}, h_{i_1}\right]$ and $\left[e_{i_2}, h_{i_2}\right]$, respectively, and $e_{i_1} < h_{i_1} < e_{i_2} < h_{i_2}$. If request $i_2$ is served before $i_1$ by the $k$th AMR on its $p$th route, then we have $P\left\{X^k_{p,i_1} > h_{i_1}\right\} = 1$ and $P\left\{Y^k_{p,i_1} > h_{i_1}\right\} = 1$.*

**Proof.** Because request $i_2$ is served before $i_1$ by the $k$th AMR on its $p$th route, we have $Y^k_{p,i_1} \geq X^k_{p,i_1} > Y^k_{p,i_2}$. Since $Y^k_{p,i_2} = \max\left\{X^k_{p,i_2}, e_{i_2}\right\} \geq e_{i_2} > h_{i_1}$, it follows that

$$P\left\{X^k_{p,i_1} > h_{i_1}\right\} = P\left\{X^k_{p,i_1} > Y^k_{p,i_2} > h_{i_1}\right\} = 1 \text{ and } P\left\{Y^k_{p,i_1} > h_{i_1}\right\} = P\left\{Y^k_{p,i_1} > Y^k_{p,i_2} > h_{i_1}\right\} = 1. \square$$

**Remark:** *From Proposition 1, we find that if the time windows satisfy $e_{i_1} < h_{i_1} < e_{i_2} < h_{i_2}$, serving request $i_1$ first is preferred, otherwise the time window of request $i_1$ is broken definitely.*

**Proposition 2.** *Let the time windows of two requests $i_1$ and $i_2$ be $\left[e_{i_1}, h_{i_1}\right]$ and $\left[e_{i_2}, h_{i_2}\right]$, respectively, and $h_1 < h_2$. The arrival times of the two requests are $X_{p,i_1}^k$ and $X_{p,i_2}^k$, which follow the same distribution. Then we have $P\{Y_{p,i_1}^k > h_{i_1}\} > P\{Y_{p,i_2}^k > h_{i_2}\}$.*

**Proof.** The probability that request $i_1$ breaks its time window is

$$P\{Y_{p,i_1}^k > h_{i_1}\} = P\{\max(X_{p,i_1}^k, e_{i_1}) > h_{i_1}\} = 1 - P\{X_{p,i_1}^k \le h_{i_1}\}P\{e_{i_1} \le h_{i_1}\} = 1 - P\{X_{p,i_1}^k \le h_{i_1}\}.$$

The probability that request $i_2$ breaks its time window is

$$P\{Y_{p,i_2}^k > h_{i_2}\} = P\{\max(X_{p,i_2}^k, e_{i_2}) > h_{i_2}\} = 1 - P\{X_{p,i_2}^k \le h_{i_2}\}P\{e_{i_2} \le h_{i_2}\} = 1 - P\{X_{p,i_2}^k \le h_{i_2}\}.$$

Note that $X_{p,i_1}^k$ and $X_{p,i_2}^k$ follow the same distribution. Let the density of $X_{p,i_1}^k$ and $X_{p,i_2}^k$ be $p(x)$, then

$$P\{X_{p,i_1}^k \le h_{i_1}\} = \int_0^{h_{i_1}} p(x)dx < \int_0^{h_{i_2}} p(x)dx = P\{X_{p,i_2}^k \le h_{i_2}\}.$$

Therefore, $P\{Y_{p,i_1}^k > h_{i_1}\} > P\{Y_{p,i_2}^k > h_{i_2}\}$. □

**Remark:** *From Proposition 2 we find that if the time windows $\left[e_{i_1}, h_{i_1}\right]$ and $\left[e_{i_2}, h_{i_2}\right]$ satisfy $h_{i_1} < h_{i_2}$, prioritizing the service of request $i_1$ leads to a smaller probability of time window violation.*

To obtain solutions within a limited time, we propose three operators, namely swap*, 2-opt*, and relocation*, based on Proposition 1 and Proposition 2.

(1) The swap* operator: figure 2a illustrates a scenario where Request 5 violates its time window constraints at its current position, while the upper bound of the time window for Request 2 is looser. Consequently, Requests 2 and 5 are chosen for swapping. On the other hand, to avoid getting stuck in local optima when all requests on the current route to satisfy their time window constraints, two requests from the current solution are randomly selected for position swapping, while the remaining requests remain unchanged. Figure 2b displays an example where Requests 3 and 6 are selected, resulting in a new solution generated through the swap operation.

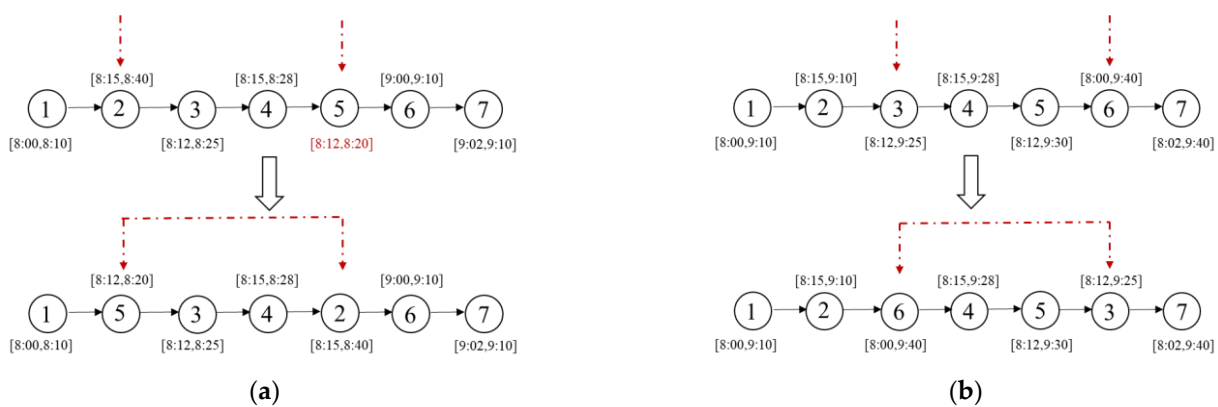

**Figure 2.** Swap* operator: (**a**) position swapping for the requests with violated time windows, (**b**) random selection for position swapping.

(2) The 2-opt* operator: as shown in Figure 3a, a declining trend is observed in the upper bound of time windows for the sub-route between Request 2 and Request 5 in the existing route. Therefore, we reverse the order of this sub-route between Requests 2 and 5 to improve the solution. For cases where there is no sub-route with decreasing time window upper bounds in the current solution, we randomly select two requests from the incumbent solution and reverse the order of points between them before inserting them back into the service route. As shown in Figure 3b, Requests 3 and 6 are randomly selected, and their routes are reversed to create a new feasible solution.

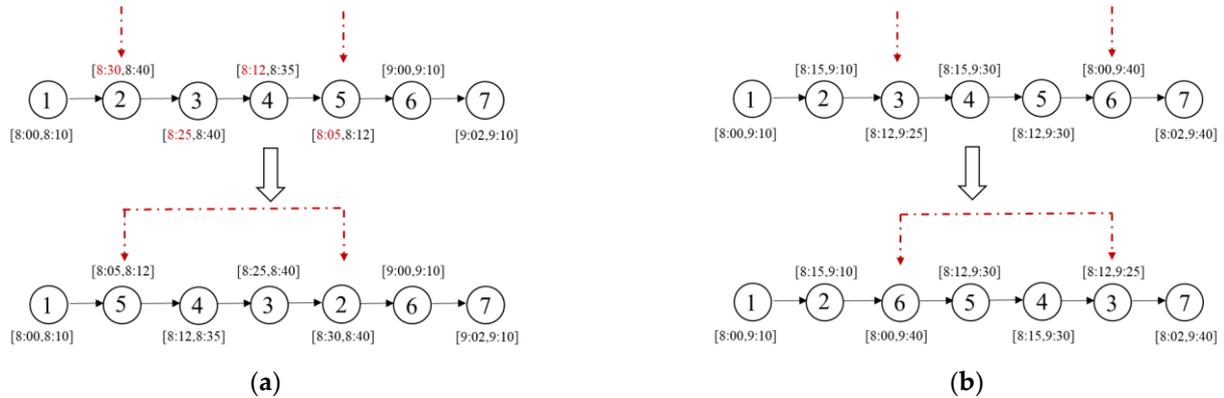

**Figure 3.** 2-opt* operator: (**a**) point reversal for the requests with violated time windows, (**b**) random selection for point reversal.

(3) The relocation* operator: Similarly, by examining the time windows of requests on the current route, as shown in Figure 4a, we notice that Request 5 exactly violates its time window constraints. Therefore, we select Request 5 and remove it from its current position, then insert it before Request 3, which has a later time window. If none of the above conditions apply to the current route, we randomly select a request, remove it from its current position, and insert it at a random position. As shown in Figure 4b, Request 6 is randomly selected and placed after Request 4 in the new solution.

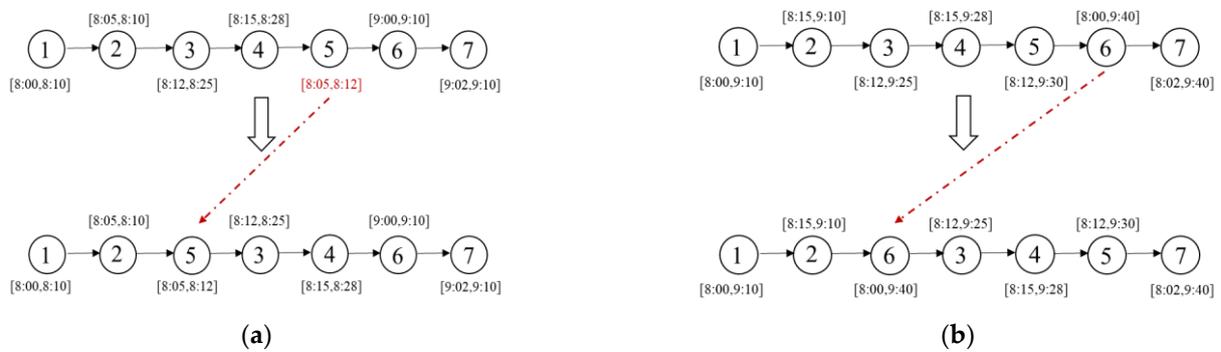

**Figure 4.** Relocation* operator: (**a**) relocation for the requests with violated time windows, (**b**) random selection for relocation.

Please note that the solutions generated by the above swap*, 2-opt*, and relocation* operations may be unfeasible as they do not consider load and battery level constraints. To address this issue, we introduce two repair operators, depot insertion and charging insertion, which can transform any infeasible solution into a feasible one. Additionally, it is important to track the number of AMRs involved in the service. By applying the AMR decrease operator to the current feasible solutions, we can minimize the participation of AMRs while still ensuring that all constraints are satisfied.

(4) Depot insertion: if a service route is not feasible due to overloading, a depot is inserted before the point where the remaining load does not meet the downward load demands. Assume that the load capacity of each AMR is 20. In Figure 5, the number above each circled node represents the demand for each request. Since $5+5+5+3+5+5 > 20$, the route $d \to 1 \to 2 \to 3 \to 4 \to 5 \to 6 \to d$ is not feasible due to overload; the first request with a negative remaining load in the current route is Request 5. The depot $d$ is inserted before Request 5, as shown in 5b. The modified feasible routes are $d \to 1 \to 2 \to 3 \to 4 \to d$, $d \to 5 \to 6 \to d$.

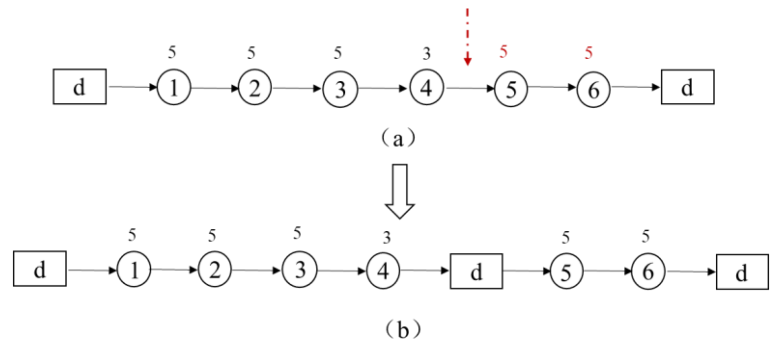

**Figure 5.** Depot insertion operator: (**a**) infeasible route violating load capacity, (**b**) feasible route after repair.

(5) Charging insertion: if a service route is not feasible due to insufficient electricity, a charging station is inserted before the point whose remaining electricity is lower than the lower bound of battery level $\alpha$. Let $\alpha = 0$. As shown in Figure 6a, route $d \to 1 \to 2 \to 3 \to 4 \to \cdots$ is not feasible due to insufficient electricity, where the number above each circled node represents the remaining battery level of the AMR. The first request with non-positive remaining electricity in the current route is Request 5. The charging station $c$ is inserted before Request 5, as shown in 5b. The modified feasible route is $d \to 1 \to 2 \to 3 \to 4 \to c \to 5 \to 6 \to d$.

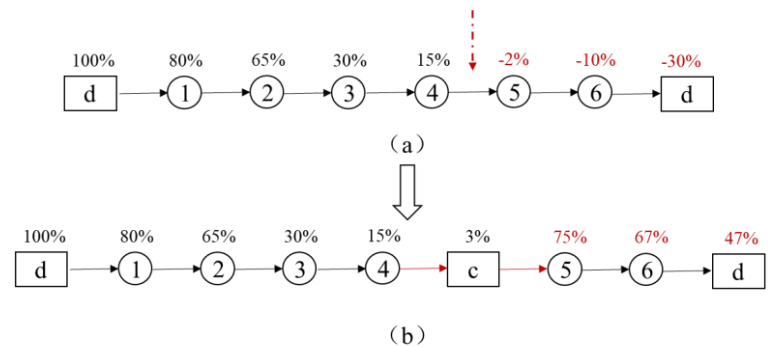

**Figure 6.** Charging insertion operator: (**a**) infeasible route violating electricity constraints, (**b**) feasible route after repair.

**Proposition 3.** *If the time windows and the battery level constraints are satisfied, reducing the number of AMRs participating in service by combining the paths of different AMRs can reduce the total cost.*

**Proof.** Assume that the requests are served by $m$ AMRs and the solution is $x = \left\{ x_{pij}^k \mid x_{pij}^k = 0, 1 \right\}$. The corresponding objective value is $f(m,x) = \xi_1 m + \xi_2 \sum_{k=1}^{m} \sum_{p \in L_k} \sum_{\substack{i \in V_1 \\ j \in V_2 \\ i \neq j}} d_{ij} x_{pij}^k$.

If the time windows and the battery level constraints are satisfied, assume that the paths of two AMRs can be completed by a single AMR. The new solution is denoted by $\tilde{x} = \left\{ \tilde{x}_{pij}^k \mid \tilde{x}_{pij}^k = 0, 1 \right\}$. We have $f(m-1, \tilde{x}) = \xi_1(m-1) + \xi_2 \sum_{k=1}^{m-1} \sum_{p \in L_k} \sum_{\substack{i \in V_1 \\ j \in V_2 \\ i \neq j}} d_{ij} \tilde{x}_{pij}^k$. Note that the combination of paths does not change the travel distance, that is

$$\sum_{k=1}^{m}\sum_{p\in L_k}\sum_{\substack{i\in V_1\\j\in V_2\\i\neq j}}d_{ij}x_{pij}^k = \sum_{k=1}^{m-1}\sum_{p\in L_k}\sum_{\substack{i\in V_1\\j\in V_2\\i\neq j}}d_{ij}\tilde{x}_{pij}^k$$, but the cost of AMRs is reduced since the AMRs are reduced in number. Therefore, $f(m-1,\tilde{x}) < f(m,x)$.

By Proposition 3, we propose the following AMR decrease operator.

(6) AMR decrease: the primary objective of this operator is to reduce the number of currently engaged AMRs in service provision while ensuring compliance with the constraints related to time windows and battery capacity. In Figure 7a, the current set of routes consists of three AMRs serving seven requests. The numbers above each request point indicate the remaining battery capacity to reach that request, while the numbers below represent the corresponding time window. Upon observing that the first AMR completes its service and returns to the depot at 8:36, and that it only takes 5 min to travel from the depot to Request 5, we conclude that the first AMR can proceed to Request 5 after completing its transportation task of Request 4. As shown in 7b, the improved set of routes becomes:

$$m = 2, x = \left\{x_{1,d,1}^1 = x_{1,1,4}^1 = x_{1,4,d}^1 = x_{2,d,5}^1 = x_{2,5,d}^1 = x_{1,d,3}^2 = x_{1,3,2}^2 = x_{1,2,7}^2 = x_{1,7,6}^2 = x_{1,6,d}^2 = 1\right\}.$$

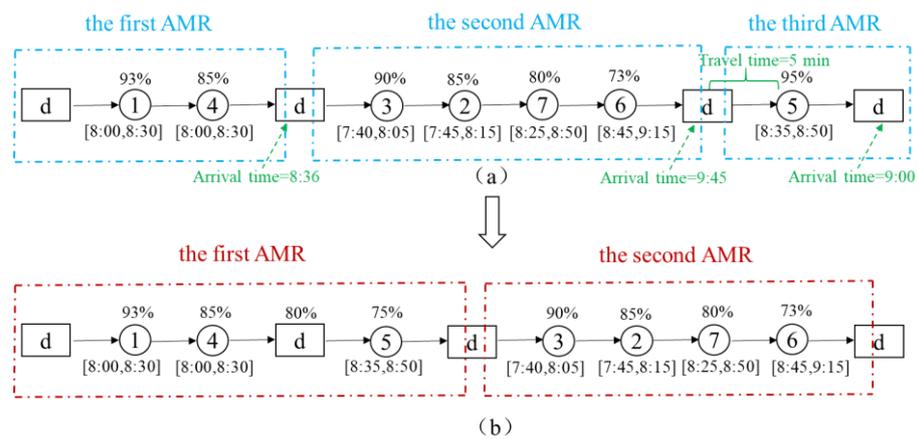

**Figure 7.** AMR decrease operator: (**a**) the set of routes before applying the AMR decrease operator, (**b**) the set of routes using the AMR decrease operator.

Although the local search strategies could achieve transformation and improvement, the solution may still fall into the local optimum. To address this issue, we use a shaking procedure that creates significant changes in the current solution. Specifically, we employ a 2-opt-L* operator from [38], and accept the new solution generated by the disturbance only if its cost function value $f(x_l')$ is less than $f(x_l)\times\delta$, where $\delta$ is the disturbance parameter and $\delta > 1$. This ensures that the new solution can jump out of the local optimal region. Algorithm 3 lines 1–8 show the details of this process.

**Algorithm 3.** Shaking Procedure

**Input**: solution $x_l$

**Output**: the shake solution $x_s$

1. **For** solution $x_l$ **do**

2. $x_l' \leftarrow \min\{\text{2-opt-L*}(x_l)\}$;//Choose the best solution from the neighborhood solutions 2-opt-L*$(x_l)$ generated by the 2-opt L* operator.

3.  **If** $f(x'_l) < f(x_l) \times \delta$ **then**// $\delta > 1$
4.       $x_s \leftarrow x'_l$
5.  **Else**
6.       $x_s \leftarrow x_l$
7.  **End if**
8. **End for**
9. **Return** $x_s$

Lines 1–6 in Algorithm 4 provide a detailed description of a feasible operation. Specifically, the depot insertion operator and charging insertion operator are employed to improve infeasible neighborhood solutions. Furthermore, the AMR decrease operator is used on Line 6 to optimize feasible solutions.

**Algorithm 4.** Feasible Operation

**Input:** solution $x$

**Output:** feasible solution $x'$

1. **If** $x$ infeasible **then**
2.    Generate a feasible solution $x''$ by depot insertion and charging insertion.
3. **Else**
4.    $x'' \leftarrow x$
5. **End**
6. Using the AMR decrease operator to optimize $x''$ and generate $x'$.
7. **Return** $x'$

## 5. Numerical Examples

This section conducts numerical experiments to verify the effectiveness of the stochastic programming model and the computation method. Section 5.1 provides parameter tuning. The effectiveness of the VNS algorithm is verified in Section 5.2. A real case study is presented in Section 5.3.

*5.1. Parameter Tuning*

Suppose that the daily fixed cost of each AMR is $\xi_1 = 30$ (unit: \$) and the travel cost per unit distance of each AMR is $\xi_2 = 0.01$ (unit: \$/m). Let $\alpha = 0$, $\beta = 0.8$. The variables $S_i$ (service time) and $T_{ij}$ (travel time) follow normal distributions and $S_i \sim N(600, \sigma_s^2)$, $T_{ij} \sim N(\mu_t, \sigma_t^2)$, where $\mu_t = d_{ij}/v_r + 6 + 51.25 \cdot I_{ij}$, $\sigma_t^2 = \sigma_0^2 + \sigma_f^2 \cdot I_{ij}$, and $I_{ij}$ is an indicator function of $|f_{ij}|$, i.e.,

$$I_{ij} = \begin{cases} 1, & |f_{ij}| \neq 0, \\ 0, & |f_{ij}| = 0. \end{cases}$$

To guarantee the service quality, the probabilistic constraint in (10) is satisfied with a confidence level of 0.95, i.e., $\varepsilon = 0.05$. The specifications of AMRs are shown in Table 2, where $r = \frac{1}{216} percent/s$ and $v_q = \frac{1}{162} percent/s$. The AMR can work for 6 h continuously, and the full charging time is 4.5 h. Without being specified, the parameters are the same in what follows.

**Table 2.** Specifications of AMRs.

| Parameter | Specification |
|---|---|
| Running Speed $v_r$ | $1\ m/s$ |
| Full Battery Capacity Level | 1 |
| Electricity Consumption Rate $r$ | $1/216\ percent/s$ |
| Charging Rate $v_q$ | $1/162\ percent/s$ |

All the following computation is run by MATLAB 2014a in the experimental environment as Inter(R) Core (TM)i9-9900K CPU @ 3.60 GHz, 16.00 GB and the operating system is Windows 10.

It is well known that CPLEX as a solver for optimization problems is suitable for solving optimization problems with small data. In this study, we compare the performance of the VNS algorithm with CPLEX. Specifically, we aim to determine the appropriate value for the iteration count $N$ in the VNS algorithm. We conducted experiments using a dataset provided by Singapore Changi General Hospital, which consists of 12 medical requests. There is a depot $d$ and a charging station $c$, and the AMR capacity $Q = 20kg$. The distance and floor difference between the 12 medical requests is presented in Tables 3 and 4, respectively. The detailed information about the demand and time window of each request can be found in Table 5.

The optimal solution obtained by CPLEX is: two AMRs are needed to run three routes, where the first AMR runs two routes $d \rightarrow 1 \rightarrow 3 \rightarrow 6 \rightarrow 7 \rightarrow d$, $d \rightarrow 9 \rightarrow 11 \rightarrow 10 \rightarrow d$; the second AMR runs the route $d \rightarrow 4 \rightarrow 2 \rightarrow 5 \rightarrow 8 \rightarrow 12 \rightarrow d$. The total cost is $f(x) = \xi_1 \cdot m + \xi_2 \cdot \sum d_{ij} = 30 \times 2 + 0.01 \times 1190 = 71.9$.

**Table 3.** Distance between the 12 medical requests.

| Distance | Depot | Requests | | | | | | | | | | | | Charging Station |
|---|---|---|---|---|---|---|---|---|---|---|---|---|---|---|
| $d_{ij}$ | d | 1 | 2 | 3 | 4 | 5 | 6 | 7 | 8 | 9 | 10 | 11 | 12 | c |
| d | 0 | 100 | 150 | 110 | 100 | 120 | 110 | 110 | 120 | 100 | 110 | 150 | 100 | 0 |
| 1 | 100 | 0 | 120 | 80 | 110 | 100 | 90 | 90 | 100 | 110 | 80 | 120 | 0 | 100 |
| 2 | 150 | 120 | 0 | 80 | 80 | 70 | 100 | 100 | 70 | 80 | 80 | 0 | 120 | 150 |
| 3 | 110 | 80 | 80 | 0 | 100 | 70 | 80 | 80 | 70 | 100 | 0 | 80 | 80 | 110 |
| 4 | 100 | 110 | 80 | 100 | 0 | 120 | 110 | 110 | 120 | 0 | 100 | 80 | 110 | 100 |
| 5 | 120 | 100 | 70 | 70 | 120 | 0 | 100 | 100 | 0 | 120 | 70 | 70 | 100 | 120 |
| 6 | 110 | 90 | 100 | 80 | 110 | 100 | 0 | 0 | 100 | 110 | 80 | 100 | 90 | 110 |
| 7 | 110 | 90 | 100 | 80 | 110 | 100 | 0 | 0 | 100 | 110 | 80 | 100 | 90 | 110 |
| 8 | 120 | 100 | 70 | 70 | 120 | 0 | 100 | 100 | 0 | 120 | 70 | 70 | 100 | 120 |
| 9 | 100 | 110 | 80 | 100 | 0 | 120 | 110 | 110 | 120 | 0 | 100 | 80 | 110 | 100 |
| 10 | 110 | 80 | 80 | 0 | 100 | 70 | 80 | 80 | 70 | 100 | 0 | 80 | 80 | 110 |
| 11 | 150 | 120 | 0 | 80 | 80 | 70 | 100 | 100 | 70 | 80 | 80 | 0 | 120 | 150 |
| 12 | 100 | 0 | 120 | 80 | 110 | 100 | 90 | 90 | 100 | 110 | 80 | 120 | 0 | 100 |
| c | 0 | 100 | 150 | 110 | 100 | 120 | 110 | 110 | 120 | 100 | 110 | 150 | 100 | 0 |

**Table 4.** Floor difference between the 12 medical requests.

| Floor Depot | | | | | | Requests | | | | | | | | Charging Station |
|---|---|---|---|---|---|---|---|---|---|---|---|---|---|---|
| $\|f_{ij}\|$ | $d$ | 1 | 2 | 3 | 4 | 5 | 6 | 7 | 8 | 9 | 10 | 11 | 12 | $c$ |
| $d$ | 0 | 1 | 6 | 2 | 5 | 3 | 4 | 4 | 3 | 5 | 2 | 6 | 1 | 0 |
| 1 | 1 | 0 | 5 | 1 | 4 | 2 | 3 | 3 | 2 | 4 | 1 | 5 | 0 | 1 |
| 2 | 6 | 5 | 0 | 4 | 1 | 3 | 2 | 2 | 3 | 1 | 4 | 0 | 5 | 6 |
| 3 | 2 | 1 | 4 | 0 | 3 | 1 | 2 | 2 | 1 | 3 | 0 | 4 | 1 | 2 |
| 4 | 5 | 4 | 1 | 3 | 0 | 2 | 1 | 1 | 2 | 0 | 3 | 1 | 4 | 5 |
| 5 | 3 | 2 | 3 | 1 | 2 | 0 | 1 | 1 | 0 | 2 | 1 | 3 | 2 | 3 |
| 6 | 4 | 3 | 2 | 2 | 1 | 1 | 0 | 0 | 1 | 1 | 2 | 2 | 3 | 4 |
| 7 | 4 | 3 | 2 | 2 | 1 | 1 | 0 | 0 | 1 | 1 | 2 | 2 | 3 | 4 |
| 8 | 3 | 2 | 3 | 1 | 2 | 0 | 1 | 1 | 0 | 2 | 1 | 3 | 2 | 3 |
| 9 | 5 | 4 | 1 | 3 | 0 | 2 | 1 | 1 | 2 | 0 | 3 | 1 | 4 | 5 |
| 10 | 2 | 1 | 4 | 0 | 3 | 1 | 2 | 2 | 1 | 3 | 0 | 4 | 1 | 2 |
| 11 | 6 | 5 | 0 | 4 | 1 | 3 | 2 | 2 | 3 | 1 | 4 | 0 | 5 | 6 |
| 12 | 1 | 0 | 5 | 1 | 4 | 2 | 3 | 3 | 2 | 4 | 1 | 5 | 0 | 1 |
| $c$ | 0 | 1 | 6 | 2 | 5 | 3 | 4 | 4 | 3 | 5 | 2 | 6 | 1 | 0 |

**Table 5.** Demand and time window of the 12 medical requests.

| Request | Demand | Time Window |
|---|---|---|
| 1 | 4 | [8:10, 8:20] |
| 2 | 4 | [8:10, 8:20] |
| 3 | 4 | [8:10, 8:20] |
| 4 | 4 | [8:10, 8:20] |
| 5 | 4 | [8:40, 8:50] |
| 6 | 4 | [8:40, 8:50] |
| 7 | 4 | [10:10, 11:00] |
| 8 | 4 | [10:10, 11:00] |
| 9 | 4 | [10:40, 11:00] |
| 10 | 4 | [10:40, 11:00] |
| 11 | 4 | [10:40, 11:00] |
| 12 | 4 | [10:40, 11:00] |

For the VNS algorithm, let the maximum number of iterations be $N = 800, 1000, 2000, 4000, 5000$. We run ten times for any fixed $N$ and get the total number of AMRs participating in service $m$, the total travel distance $\sum d_{ij}$, and the total cost $f(x)$. The optimal solution of the VNS algorithm for any fixed $N$ is the one with the least cost among the ten experiments, and the results are shown in Table 6. As $N$ increases, $f(x)$ approaches the optimal value 71.9, which is the same as the result of CPLEX.

**Table 6.** The results of the VNS algorithm.

| $N$ | $m$ | $\sum d_{ij}$ | $f(x)$ | Time (s) |
|---|---|---|---|---|
| 800 | 3 | 1100 | 101.00 | 2.17 |
| 1000 | 2 | 1230 | 72.30 | 2.36 |
| 2000 | 2 | 1220 | 72.20 | 2.68 |
| 4000 | 2 | 1190 | 71.90 | 2.93 |
| 5000 | 2 | 1190 | 71.90 | 3.10 |

It can be found from Table 6 that when the number of iterations is $N=800$, the total travel distance $\sum d_{ij}$ is the shortest, but the number of AMRs participating in service is more than the others. The cost $f(x)$ with three AMRs participating in service is much more than the cost with two AMRs participating in the service, even though the travel distance is short. It is a better choice to adopt fewer AMRs to provide service. The optimal solution is described in detail with a route diagram, as shown in Figure 8, where ① represents request $i$, and the three elements in vector $(l_i\ level,\ \gamma_i\ \%,\ u_i\ kg)$ represent the floor, the remaining electricity, and the remaining load at the request point $i$, respectively, $i=1,\cdots,12$.

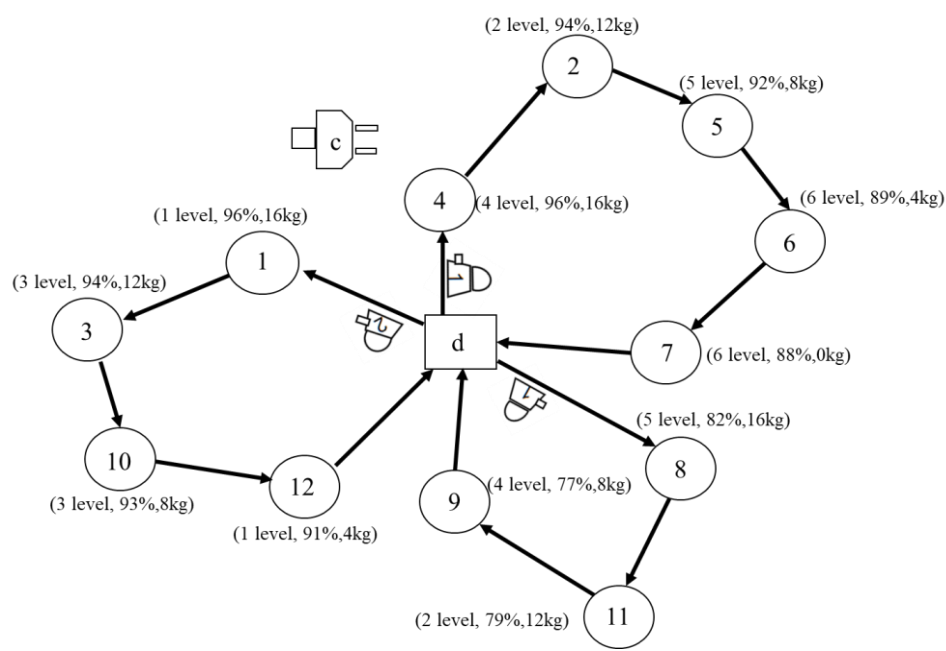

**Figure 8.** Optimal service routes of medical AMRs with 12 requests.

*5.2. Comparative Experiments*

In this section, three comparative experiments are conducted. Firstly, the performance of the proposed methodology is compared with CPLEX in Section 5.2.1. Secondly, the influence of the variance of random variables is investigated in Section 5.2.2. Finally, we study the sensitivity of the optimal solution to the travel distance in Section 5.2.3.

5.2.1. Experiment 1

To verify the feasibility and effectiveness of the algorithm, computational experiments are conducted by six modified Solomon instances. The tabu search (TS) algorithm is a heuristic algorithm based on neighborhood search proposed by Glover [39]. It can solve a class of combinatorial optimization problems including variants of the VRP that involve stochastic factors. For small instances, the performance of the VNS algorithm is compared with CPLEX. For large instances, we employed a tabu search algorithm (TS) and VNS algorithm for experimental comparison. In the case of small instances, we select the first 15 nodes from these Solomon instances. The coordinates, demands, and time windows of the nodes remain unchanged. Each node is assigned a floor information where the first five requests are located on the first floor, the sixth to tenth requests on the second floor, and so on. The AMR's capacity is the original vehicle capacity of the Solomon instance, while the other parameters remain unchanged, as shown in Table 2.

Table 7 presents a comparative analysis of the CPLEX optimizer and the VNS algorithm to solve six modified small Solomon instances. We set the maximum runtime per event to 1800 s for CPLEX. The iteration number of the VNS algorithm is 5000, and we run ten times for each instance. Here, $m$ represents the total number of AMRs, $\sum d_{ij}$ denotes the sum of travel distance, $Time$ indicates the total running time (unit: second), and $f$ is the objective value. For the VNS, we present both the best (*Best*) and average (*Avg*) values for the following metrics: the number of AMRs, travel distance, running time, and objective value. We define $Gap = (f - f^*)/f * 100$ and $G' = (\bar{f} - f^*)/\bar{f} * 100$, where $f^*$ is the best objective function value obtained by VNS, $\bar{f}$ is the average objective function value obtained by VNS, and $f$ is the best lower bound of the objective function of the final solution obtained by the CPLEX.

**Table 7.** Comparison between the CPLEX and the VNS algorithms for the small instances.

| Instance | CPLEX | | | | VNS | | | | | | $f = \xi_1 m + \xi_2 \sum d_{ij}$ | | | Gap (%) |
| | | | | | $m$ | | $\sum d_{ij}$ | | Time (s) | | | | | |
| | $m$ | $\sum d_{ij}$ | Time (s) | $f$ | Best | Avg | Best | Avg | Best | Avg | $f^*$ | $\bar{f}$ | $G'$ | |
| --- | --- | --- | --- | --- | --- | --- | --- | --- | --- | --- | --- | --- | --- | --- |
| R101-15 | 12 | 248.91 | 1800 | 362.48 | 12 | 12 | 603 | 611.0 | 31.06 | 32.03 | 366.02 | 366.11 | 0.02 | −0.66 |
| R201-15 | 3 | 327.35 | 12.53 | 93.27 | 3 | 3.6 | 397.4 | 378.1 | 24.59 | 25.05 | 93.97 | 111.78 | 15.9 | −0.75 |
| C101-15 | 4 | 104.80 | 1800 | 121.04 | 4 | 4.6 | 220 | 270.1 | 25.37 | 26.17 | 122 | 140.7 | 13.2 | −0.95 |
| C201-15 | 2 | 189.26 | 8.47 | 61.89 | 2 | 2.6 | 211.2 | 245.3 | 23.16 | 24.84 | 62.11 | 80.45 | 22.7 | −0.35 |
| RC102-15 | 12 | 186.94 | 1800 | 361.86 | 12 | 12.2 | 868.2 | 881.1 | 30.69 | 31.44 | 368.6 | 374.8 | 1.65 | −1.88 |
| RC201-15 | 3 | 198.30 | 12.51 | 91.98 | 3 | 3.2 | 310 | 350 | 22.42 | 23.97 | 93.10 | 99.50 | 6.43 | −1.22 |
| Average | | | | | | | ----- | | | | | | 10.0 | −0.96 |

Table 7 shows that although the objective function value obtained through VNS does not reach the optimality achieved by CPLEX, the difference was less than 1% and the average gap (%) is −0.96. This indicates the effectiveness of the VNS algorithm in finding near-optimal solutions. The VNS algorithm achieved competitive performance compared to CPLEX in terms of optimizing the objective function.

We compare the VNS algorithm with ten repeated runs on different instances, and the results are shown in Table 7, where $f^*$ and $\bar{f}$ are the optimal and average values of the ten runs, respectively. The smaller the absolute value of $G'$, the better the robustness of the VNS algorithm. Table 7 shows that the average value of $G'$ is about 10%. Overall, these results highlight the effectiveness and robustness of the VNS algorithm.

Table 8 provides a comparative analysis between the TS algorithm and the VNS algorithm for solving six modified large Solomon instances. It is noted that each experiment has been executed 10 times and the best value per set of 10 executions is recorded. The table includes the number of AMRs ($m$), travel distance ($\sum d_{ij}$), and objective value, where $f_{VNS}$ and $f_{TS}$ are the best objective values obtained by VNS and TS, respectively. Let $G_1 = (f_{VNS} - f_{TS})/f_{VNS} * 100$. If $G_1 < 0$, the VNS algorithm performs better. The running time of both algorithms is set to be equal.

**Table 8.** Comparison between the TS and the VNS algorithms for the large instances.

| Instance | TS | VNS |
| --- | --- | --- |

|      | $m$ | $\sum d_{ij}$ | $f_{TS}$ | $m$ | $\sum d_{ij}$ | $f_{VNS}$ | $G_1$ |
|------|-----|---------------|----------|-----|---------------|-----------|-------|
| R101 | 87 | 121.92 | 2731.93 | 87 | 122.2 | 2732.2 | 0.01 |
| R201 | 47 | 82.791 | 1492.79 | 47 | 83.59 | 1493.6 | 0.05 |
| C101 | 62 | 148.6 | 2008.6 | 62 | 145.3 | 2005.3 | −0.16 |
| C201 | 37 | 109.8 | 1219.8 | 35 | 118.6 | 1168.6 | −4.38 |
| RC102 | 87 | 148.31 | 2758.31 | 86 | 147.1 | 2727.1 | −1.14 |
| RC201 | 51 | 108.9 | 1638.9 | 50 | 107.8 | 1607.8 | −1.93 |
| Average | 61.83 | 120.05 | 1975.1 | 61.2 | 120.7 | 1955.8 | −1.26 |

The experimental results presented in Table 8 reveal a comparison between the TS algorithm and the VNS algorithm. The findings indicate that the VNS algorithm generates a higher quality solution than the TS algorithm ($G_1 = -1.26\%$).

It is worth noting that if the distribution of request nodes is relatively random, such as R101 and R201, the VNS algorithm obtains the optimal solution that is inferior to the TS algorithm. However, for the remaining four instances, the VNS algorithm generates higher quality solutions than TS algorithm. The results of this experiment verify that the VNS algorithm is effective.

5.2.2. Experiment 2

In Experiment 2, we analyze the influence of the variance of variables by setting $\sigma_f^2 = 16n$, $\sigma_0^2 = 4n$, $\sigma_s^2 = 36n$, $n = 1, 10, 100$. The variances of variables $T_{ij}$ and $S_i$ increase with the value of $n$. All the other parameters are the same as Experiment 1.

We employ the VNS algorithm in Section 4 to solve this problem. This algorithm calculates the number of participating AMRs, travel distance, and total cost. Table 9 demonstrates that when $n = 1$ or 10, only two AMRs are involved in the service. However, for $n = 100$, four AMRs are required. The increasing variance of service and travel times results in a higher probability of violating time windows. To avoid the violation of time windows, increasing the number of AMRs is an effective approach. Consequently, the total cost rises from 71.9 to 131.8 due to the increased cost of additional AMRs.

**Table 9.** The effect of variance of variables on the AMR scheduling problem.

| $n$ | $m$ | $\sum d_{ij}$ | $f$ |
|-----|-----|---------------|-----|
| 1 | 2 | 1190 | 71.9 |
| 10 | 2 | 1190 | 71.9 |
| 100 | 4 | 1180 | 131.8 |

This discovery indicates that the optimal number of AMRs depends on the numerical characteristics of service and travel time, and the system can better adapt to the randomness of service and travel time by allocating more AMRs. Correspondingly, the probability of violating time windows is reduced. By understanding how variables affect system performance, decision-makers can make wise choices to improve operational efficiency and minimize interference caused by time window violations.

5.2.3. Experiment 3

In Sections 5.1., 5.2.1., and 5.2.2., the electricity consumption is low and the AMRs do not visit the charging station due to sufficient remaining electricity. In Experiment 3, we investigate the impact of distance on charging times and the number of AMRs. Let the travel distance be $nd_{ij}$, where $n = 1, 2, 4, 6, 8$ and $d_{ij}$ represents the distance between $i$ and $j$ given in Table 3 for $i, j = 1, \cdots, 12$. The variances of $T_{ij}$ and $S_i$ are denoted by $\sigma_t^2 = 40 + 160 I_{ij}$ and $\sigma_s^2 = 360$, respectively. All the other parameters are the same as those in Section 5.1.

Note that long distance results in long travel time and increased power consumption. As a result, the number of AMRs engaged in service, or the frequency of charging times, shows a corresponding increase with the progressive increment of travel distance as shown in Table 10. These findings have practical significance for system design and optimization. Understanding the relationship between travel distance, power consumption, and the number of AMRs, decision-makers can make wise decisions to improve operational efficiency and optimize resource utilization in real-world applications.

Table 10. The influence of distance on charging times and total number of AMR.

| Distance | $d_{ij}$ | $2d_{ij}$ | $4d_{ij}$ | $6d_{ij}$ | $8d_{ij}$ |
|---|---|---|---|---|---|
| Number of AMRs | 2 | 2 | 4 | 4 | 8 |
| Charging Times | 0 | 0 | 0 | 0 | 1 |

*5.3. Case Study*

As mentioned in the introduction, this study was provoked by the actual demands of Singapore Changi General Hospital using AMRs to perform some service. This section addresses the realistic problem encountered at the hospital with the approach proposed in this paper.

A set of homogeneous AMRs is utilized to fulfill specific medical requests at the hospital. These requests involve delivering drugs from a pharmacy to 32 wards twice daily. The 64 recorded requests are denoted by $1, 2, \cdots, 64$, where requests $i$ and $32 + i$ belong to the same ward. The depot, denoted by 0, serves as both the starting point and the location for charging the AMRs. Details regarding the demand, mean service time, and time window of these 64 requests can be found in Table A1 of the Appendix A. Additionally, Tables A2 and A3 are the distance and floor difference between the 32 wards, respectively. Let $\sigma_t^2 = 40 + 160 I_{ij}$, $\sigma_s^2 = 360$ and the initial battery level of the AMRs be 50%, while the remaining parameters are the same as those in Section 5.1.

Table 11 presents the optimized service routes, where the first AMR runs three routes, the second AMR runs three routes, and the third AMR runs two routes. The computed metrics for each route include travel distance, total load, electricity consumption, and the mean arrival time of requests. For instance, the first route of the third AMR is $0 \to 45 \to 13 \to 53 \to 21 \to 17 \to 49 \to 41 \to 31 \to 63 \to c \to 0$. This route covers a distance of 926 m with a total load of 18 kg and its charge time is one. The mean arrival time for the requests in this route is sequentially recorded as 11:00:00-11:06:10-11:11:10-11:25:39-11:30:39-11:38:54-11:43:54-11:51:59-12:04:03-12:09:03-12:17:20-12:22:20. In the entire service process, each AMR is charged once.

Via the case study, the effectiveness and feasibility of the VNS algorithm solving the routing problem of AMRs at smart hospitals are verified.

Table 11. Optimal service routes of AMRs for drug delivery.

| AMR No. | Service Route | Distance (m) | Load (kg) | Number of Charges | Mean Arrival Time of Requests |
|---|---|---|---|---|---|
| 1 | ① $0 \to 61 \to 29 \to 27 \to 59 \to 30 \to 62 \to 26 \to 58 \to 0$ | 506 | 16 | 0 | 10:20:00-10:30:35-10:55:50-11:20:22-11:25:22-→ 11:32:56-11:37:56-11:45:14-11:50:14-12:00:31 |
| | ② $0 \to 5 \to 37 \to 56 \to 24 \to 64 \to 7 \to 39 \to 38 \to 51 \to 19 \to 0$ | 692 | 20 | 0 | 12:05:31-12:09:53-12:14:53-12:22:50-12:27:50-12:36:23-12:42:47-12:47:47-12:55:45-13:03:52-13:08:52-13:18:24 |
| | ③ $0 \to c \to 32 \to 33 \to 1 \to 55 \to 23 \to 0$ | 474 | 10 | 1 | 13:23:24-13:28:24-13:47:55-13:55:54-14:00:54-14:08:59-14:13:59-14:23:41 |
| 2 | ① $0 \to 60 \to 28 \to 57 \to 25 \to 0$ | 437 | 8 | 0 | 10:35:00-10:50:23-10:55:23-11:02:46-11:07:46-11:18:01 |
| | ② $0 \to c \to 9 \to 10 \to 15 \to 47 \to 12 \to 44 \to 16 \to 48 \to 11 \to 43 \to 0$ | 604 | 20 | 1 | 11:23:01-11:33:01-11:40:29-11:47:14-11:55:14-12:00:14-12:08:19-12:13:19-12:21:21-12:26:21-12:34:21-12:39:21-12:46:49 |

| | | | | |
|---|---|---|---|---|
| 3 | ③ 0 → 42 → 14 → 46 → 20 → 52 → 18 → 50 → 6 → 34 → 2 → 0 | 670 | 20 | 0 | 12:56:49-13:04:17-13:12:20-13:17:20-13:23:50-13:28:50-13:37:10-13:42:10-13:48:20-13:56:05-14:01:05-14:10:20 |
| | ① 0 → 45 → 13 → 53 → 21 → 17 → 49 → 41 → 31 → 63 → c → 0 | 926 | 18 | 1 | 11:00:00-11:06:10-11:11:10-11:25:39-11:30:39-11:38:54-11:43:54-11:51:59-12:04:03-12:09:03-12:17:20-12:22:20 |
| | ② 0 → 40 → 8 → 4 → 36 → 3 → 35 → 22 → 54 → 0 | 498 | 16 | 0 | 12:27:20-12:27:20-12:34:48-12:39:48-12:47:34-12:52:34-13:00:11-13:05:11-13:13:08-13:18:08-13:25:49-13:30:49 |

## 6. Conclusions

This paper addresses the route scheduling problem for medical AMRs at hospitals. The study focuses on optimizing the total daily cost of the hospital by minimizing the number of AMRs and travel distance. We develop a stochastic programming model and propose an adjusted variable neighborhood search algorithm to accomplish this objective. The experimental findings illustrate that appropriately organizing the driving routes of AMRs for charging and service requests can considerably decrease the hospital's overall cost while fulfilling medical requirements.

Note that in some practical scenarios, demands may arise dynamically, such as garbage collection in wards, and stochastic load demands may arise during the service process. In such cases, AMRs need to adjust their service routes based on new arrival requests. Future research can extend the AMR scheduling problem to study the optimization of routes with stochastic load demands and dynamic requests.

**Author Contributions:** Conceptualization, L.-L.C., N.Z., K.W., and Z.-B.C.; methodology, L.-L.C., N.Z., K.W., and Z.-B.C.; software, L.-L.C.; validation, L.-L.C.; writing—original draft preparation, L.-L.C.; writing—review and editing, L.-L.C., N.Z., and K.W.; visualization, L.-L.C.; supervision, L.-L.C., and N.Z. All authors have read and agreed to the published version of the manuscript.

**Funding:** This research was supported in part by State Key Laboratory of Industrial Control Technology (Grant No. ICT2021B51).

**Institutional Review Board Statement:** Not applicable.

**Informed Consent Statement:** Not applicable.

**Data Availability Statement:** The data used to support the findings of this paper are available from the corresponding author upon request.

**Conflicts of Interest:** The authors declare no conflict of interest.

## Appendix A

**Table A1.** Drug delivery requests.

| Request | Demand $q_i$ | Service Time $E(S_i)$ | Time Window $[e_i, h_i]$ |
|---|---|---|---|
| 1 | 2 | 300 | [11:21,14:23] |
| 2 | 2 | 300 | [11:24,14:18] |
| 3 | 2 | 300 | [11:09,14:02] |
| 4 | 2 | 300 | [11:08,14:09] |
| 5 | 2 | 300 | [11:22,14:27] |
| 6 | 2 | 300 | [11:13,14:25] |
| 7 | 2 | 300 | [10:59,14:18] |
| 8 | 2 | 300 | [10:56,14:29] |
| 9 | 2 | 300 | [11:23,14:26] |
| 10 | 2 | 300 | [10:59,14:19] |
| 11 | 2 | 300 | [11:20,14:26] |
| 12 | 2 | 300 | [10:48,14:15] |
| 13 | 2 | 300 | [10:54,14:09] |
| 14 | 2 | 300 | [11:26,14:02] |
| 15 | 2 | 300 | [10:38,14:14] |
| 16 | 2 | 300 | [10:39,14:25] |
| 17 | 2 | 300 | [10:44,14:16] |
| 18 | 2 | 300 | [11:07,17:00] |
| 19 | 2 | 300 | [11:16,14:13] |
| 20 | 2 | 300 | [11:02,14:09] |
| 21 | 2 | 300 | [11:02,14:05] |
| 22 | 2 | 300 | [10:43,14:23] |
| 23 | 2 | 300 | [10:39,14:19] |
| 24 | 2 | 300 | [11:26,14:26] |
| 25 | 2 | 300 | [10:56,14:06] |
| 26 | 2 | 300 | [11:14,14:14] |
| 27 | 2 | 300 | [11:20,14:16] |
| 28 | 2 | 300 | [10:43,14:02] |
| 29 | 2 | 300 | [10:55,14:23] |
| 30 | 2 | 300 | [11:02,14:25] |
| 31 | 2 | 300 | [11:11,14:13] |
| 32 | 6 | 300 | [11:21,14:15] |
| 33 | 2 | 300 | [10:43,14:08] |
| 34 | 2 | 300 | [10:48,14:29] |
| 35 | 2 | 300 | [11:12,14:17] |
| 36 | 2 | 300 | [11:09,14:14] |
| 37 | 2 | 300 | [10:32,14:01] |
| 38 | 2 | 300 | [11:16,14:18] |
| 39 | 2 | 300 | [11:47,14:34] |
| 40 | 2 | 300 | [10:51,14:08] |
| 41 | 2 | 300 | [10:47,14:23] |
| 42 | 2 | 300 | [10:40,14:14] |
| 43 | 2 | 300 | [11:05,14:06] |
| 44 | 2 | 300 | [10:50,14:11] |
| 45 | 2 | 300 | [11:06,14:00] |
| 46 | 2 | 300 | [11:14,14:25] |
| 47 | 2 | 300 | [11:04,14:28] |
| 48 | 2 | 300 | [10:37,14:03] |
| 49 | 2 | 300 | [11:17,14:26] |
| 50 | 2 | 300 | [11:08,14:10] |
| 51 | 2 | 300 | [10:32,14:04] |
| 52 | 2 | 300 | [10:54,14:24] |
| 53 | 2 | 300 | [11:25,14:20] |
| 54 | 2 | 300 | [11:16,14:14] |
| 55 | 2 | 300 | [11:14,14:19] |
| 56 | 2 | 300 | [11:41,14:13] |
| 57 | 2 | 300 | [11:02,14:04] |
| 58 | 2 | 300 | [11:24,14:05] |
| 59 | 2 | 300 | [10:49,14:02] |
| 60 | 2 | 300 | [10:50,14:22] |
| 61 | 2 | 300 | [10:30,14:15] |
| 62 | 2 | 300 | [11:05,14:10] |
| 63 | 2 | 300 | [11:09,14:24] |
| 64 | 6 | 300 | [10:53,14:26] |

**Table A2.** Distance between the wards.

| Distance | Depot | | | | | | | Requests | | | | | | | | | | | | | | | | | | | | | | | | | |
|---|---|---|---|---|---|---|---|---|---|---|---|---|---|---|---|---|---|---|---|---|---|---|---|---|---|---|---|---|---|---|---|---|---|
| $d_{ij}$ | 0 | 1 | 2 | 3 | 4 | 5 | 6 | 7 | 8 | 9 | 10 | 11 | 12 | 13 | 14 | 15 | 16 | 17 | 18 | 19 | 20 | 21 | 22 | 23 | 24 | 25 | 26 | 27 | 28 | 29 | 30 | 31 | 32 |
| 0 | 0 | 141 | 141 | 141 | 141 | 141 | 141 | 148 | 148 | 148 | 148 | 148 | 148 | 157 | 157 | 157 | 157 | 157 | 157 | 161 | 161 | 161 | 161 | 161 | 161 | 201 | 201 | 201 | 201 | 200 | 197 | 197 | 163 |
| 1 | 141 | 0 | 54 | 54 | 54 | 54 | 54 | 61 | 61 | 61 | 61 | 61 | 61 | 70 | 70 | 70 | 70 | 70 | 70 | 74 | 74 | 74 | 74 | 74 | 74 | 311 | 311 | 311 | 310 | 310 | 307 | 307 | 76 |
| 2 | 141 | 54 | 0 | 54 | 54 | 54 | 54 | 61 | 61 | 61 | 61 | 61 | 61 | 70 | 70 | 70 | 70 | 70 | 70 | 74 | 74 | 74 | 74 | 74 | 74 | 311 | 311 | 311 | 310 | 310 | 307 | 307 | 76 |
| 3 | 141 | 54 | 54 | 0 | 54 | 54 | 54 | 61 | 61 | 61 | 61 | 61 | 61 | 70 | 70 | 70 | 70 | 70 | 70 | 74 | 74 | 74 | 74 | 74 | 74 | 311 | 311 | 311 | 310 | 310 | 307 | 307 | 76 |
| 4 | 141 | 54 | 54 | 54 | 0 | 54 | 54 | 61 | 61 | 61 | 61 | 61 | 61 | 70 | 70 | 70 | 70 | 70 | 70 | 74 | 74 | 74 | 74 | 74 | 74 | 311 | 311 | 311 | 310 | 310 | 307 | 307 | 76 |
| 5 | 141 | 54 | 54 | 54 | 54 | 0 | 54 | 61 | 61 | 61 | 61 | 61 | 61 | 70 | 70 | 70 | 70 | 70 | 70 | 74 | 74 | 74 | 74 | 74 | 74 | 311 | 311 | 311 | 310 | 310 | 307 | 307 | 76 |
| 6 | 141 | 54 | 54 | 54 | 54 | 54 | 0 | 61 | 61 | 61 | 61 | 61 | 61 | 70 | 70 | 70 | 70 | 70 | 70 | 74 | 74 | 74 | 74 | 74 | 74 | 311 | 311 | 311 | 310 | 310 | 307 | 307 | 76 |
| 7 | 148 | 61 | 61 | 61 | 61 | 61 | 61 | 0 | 69 | 69 | 69 | 69 | 69 | 77 | 77 | 77 | 77 | 77 | 77 | 82 | 82 | 82 | 82 | 82 | 82 | 318 | 318 | 318 | 318 | 318 | 315 | 315 | 84 |
| 8 | 148 | 61 | 61 | 61 | 61 | 61 | 61 | 69 | 0 | 69 | 69 | 69 | 77 | 77 | 77 | 77 | 77 | 77 | 77 | 82 | 82 | 82 | 82 | 82 | 82 | 318 | 318 | 318 | 318 | 318 | 315 | 315 | 84 |
| 9 | 148 | 61 | 61 | 61 | 61 | 61 | 61 | 69 | 69 | 0 | 0 | 69 | 77 | 77 | 77 | 77 | 77 | 77 | 77 | 82 | 82 | 82 | 82 | 82 | 82 | 318 | 318 | 318 | 318 | 318 | 315 | 315 | 84 |
| 10 | 148 | 61 | 61 | 61 | 61 | 61 | 61 | 69 | 69 | 0 | 0 | 69 | 77 | 77 | 77 | 77 | 77 | 77 | 77 | 82 | 82 | 82 | 82 | 82 | 82 | 318 | 318 | 318 | 318 | 318 | 315 | 315 | 84 |
| 11 | 148 | 70 | 61 | 61 | 61 | 61 | 61 | 69 | 69 | 69 | 69 | 0 | 77 | 77 | 77 | 77 | 77 | 77 | 77 | 82 | 82 | 82 | 82 | 82 | 82 | 318 | 318 | 318 | 318 | 318 | 315 | 315 | 84 |
| 12 | 148 | 70 | 61 | 61 | 61 | 61 | 61 | 69 | 77 | 77 | 77 | 77 | 0 | 77 | 77 | 77 | 77 | 77 | 77 | 82 | 82 | 82 | 82 | 82 | 82 | 318 | 318 | 318 | 318 | 318 | 315 | 315 | 84 |
| 13 | 157 | 70 | 70 | 70 | 70 | 70 | 70 | 77 | 77 | 77 | 77 | 77 | 77 | 0 | 86 | 86 | 86 | 90 | 90 | 90 | 90 | 90 | 90 | 327 | 327 | 327 | 327 | 326 | 323 | 323 | 323 | 323 | 93 |
| 14 | 157 | 70 | 70 | 70 | 70 | 70 | 70 | 77 | 77 | 77 | 77 | 77 | 77 | 86 | 0 | 86 | 86 | 90 | 90 | 90 | 90 | 90 | 90 | 327 | 327 | 327 | 327 | 326 | 323 | 323 | 323 | 323 | 93 |
| 15 | 157 | 70 | 70 | 70 | 70 | 70 | 70 | 77 | 77 | 77 | 77 | 77 | 77 | 86 | 86 | 0 | 86 | 90 | 90 | 90 | 90 | 90 | 90 | 327 | 327 | 327 | 327 | 326 | 323 | 323 | 323 | 323 | 93 |
| 16 | 157 | 70 | 70 | 70 | 70 | 70 | 70 | 77 | 77 | 77 | 77 | 77 | 77 | 86 | 86 | 86 | 0 | 90 | 90 | 90 | 90 | 90 | 90 | 327 | 327 | 327 | 327 | 326 | 323 | 323 | 323 | 323 | 93 |
| 17 | 157 | 70 | 70 | 70 | 70 | 70 | 70 | 77 | 77 | 77 | 77 | 77 | 77 | 90 | 90 | 90 | 90 | 0 | 90 | 90 | 90 | 90 | 90 | 327 | 327 | 327 | 327 | 326 | 323 | 323 | 323 | 323 | 93 |
| 18 | 157 | 70 | 70 | 70 | 70 | 70 | 70 | 77 | 82 | 82 | 82 | 82 | 82 | 90 | 90 | 90 | 90 | 90 | 0 | 90 | 90 | 90 | 90 | 327 | 327 | 327 | 327 | 326 | 323 | 323 | 323 | 323 | 93 |
| 19 | 161 | 74 | 74 | 74 | 74 | 74 | 74 | 82 | 82 | 82 | 82 | 82 | 82 | 90 | 90 | 90 | 90 | 90 | 90 | 0 | 95 | 95 | 95 | 332 | 332 | 332 | 332 | 331 | 328 | 328 | 328 | 328 | 97 |
| 20 | 161 | 74 | 74 | 74 | 74 | 74 | 74 | 82 | 82 | 82 | 82 | 82 | 82 | 90 | 90 | 90 | 90 | 90 | 90 | 95 | 0 | 95 | 95 | 332 | 332 | 332 | 332 | 331 | 328 | 328 | 328 | 328 | 97 |
| 21 | 161 | 74 | 74 | 74 | 74 | 74 | 74 | 82 | 82 | 82 | 82 | 82 | 82 | 90 | 90 | 90 | 90 | 90 | 90 | 95 | 95 | 0 | 95 | 332 | 332 | 332 | 332 | 331 | 328 | 328 | 328 | 328 | 97 |
| 22 | 161 | 74 | 74 | 74 | 74 | 74 | 74 | 82 | 82 | 82 | 82 | 82 | 82 | 90 | 90 | 90 | 90 | 90 | 90 | 95 | 95 | 95 | 0 | 332 | 332 | 332 | 332 | 331 | 328 | 328 | 328 | 328 | 97 |
| 23 | 161 | 74 | 74 | 74 | 74 | 74 | 74 | 82 | 82 | 82 | 82 | 82 | 82 | 327 | 327 | 327 | 327 | 327 | 327 | 332 | 332 | 332 | 332 | 0 | 332 | 332 | 332 | 331 | 328 | 328 | 328 | 328 | 97 |
| 24 | 161 | 74 | 74 | 74 | 74 | 74 | 74 | 82 | 318 | 318 | 318 | 318 | 318 | 327 | 327 | 327 | 327 | 327 | 327 | 332 | 332 | 332 | 332 | 332 | 0 | 332 | 332 | 331 | 328 | 328 | 328 | 328 | 97 |
| 25 | 201 | 311 | 311 | 311 | 310 | 310 | 310 | 318 | 318 | 318 | 318 | 318 | 318 | 327 | 327 | 327 | 327 | 327 | 327 | 332 | 332 | 332 | 332 | 332 | 332 | 0 | 39 | 39 | 35 | 35 | 35 | 35 | 334 |
| 26 | 201 | 311 | 311 | 311 | 310 | 310 | 310 | 318 | 318 | 318 | 318 | 318 | 318 | 327 | 327 | 327 | 327 | 327 | 327 | 332 | 332 | 332 | 332 | 332 | 332 | 39 | 0 | 39 | 35 | 35 | 35 | 35 | 334 |
| 27 | 201 | 311 | 311 | 311 | 310 | 310 | 310 | 318 | 318 | 318 | 318 | 318 | 318 | 326 | 326 | 326 | 326 | 326 | 326 | 331 | 331 | 331 | 331 | 331 | 331 | 39 | 39 | 0 | 35 | 35 | 35 | 35 | 334 |
| 28 | 201 | 311 | 311 | 311 | 310 | 310 | 310 | 318 | 318 | 318 | 318 | 318 | 318 | 323 | 323 | 323 | 323 | 323 | 323 | 328 | 328 | 328 | 328 | 328 | 328 | 35 | 35 | 35 | 0 | 35 | 35 | 35 | 334 |
| 29 | 200 | 310 | 310 | 310 | 310 | 310 | 310 | 318 | 315 | 315 | 315 | 315 | 315 | 323 | 323 | 323 | 323 | 323 | 323 | 328 | 328 | 328 | 328 | 328 | 328 | 35 | 35 | 35 | 35 | 0 | 35 | 35 | 334 |
| 30 | 197 | 307 | 307 | 307 | 307 | 307 | 307 | 315 | 315 | 315 | 315 | 315 | 315 | 323 | 323 | 323 | 323 | 323 | 323 | 328 | 328 | 328 | 328 | 328 | 328 | 35 | 35 | 35 | 35 | 35 | 0 | 31 | 330 |
| 31 | 197 | 307 | 307 | 307 | 307 | 307 | 307 | 315 | 315 | 315 | 315 | 315 | 315 | 323 | 323 | 323 | 323 | 323 | 323 | 328 | 328 | 328 | 328 | 328 | 328 | 35 | 35 | 35 | 35 | 35 | 31 | 0 | 330 |
| 32 | 163 | 76 | 76 | 76 | 76 | 76 | 76 | 84 | 84 | 84 | 84 | 84 | 84 | 93 | 93 | 93 | 93 | 93 | 93 | 97 | 97 | 97 | 97 | 97 | 97 | 334 | 334 | 334 | 334 | 333 | 330 | 330 | 163 |

**Table A3.** Floor difference between the wards.

| Floor | Depot | | | | | | | | | | | | | | | | | | | | | | | | | | | | | | | | Requests |
|---|---|---|---|---|---|---|---|---|---|---|---|---|---|---|---|---|---|---|---|---|---|---|---|---|---|---|---|---|---|---|---|---|
| $f_{ij}$ | 0 | 1 | 2 | 3 | 4 | 5 | 6 | 7 | 8 | 9 | 10 | 11 | 12 | 13 | 14 | 15 | 16 | 17 | 18 | 19 | 20 | 21 | 22 | 23 | 24 | 25 | 26 | 27 | 28 | 29 | 30 | 31 | 32 |
| 0 | 0 | 4 | 5 | 6 | 7 | 8 | 9 | 3 | 5 | 5 | 7 | 8 | 9 | 4 | 5 | 6 | 7 | 8 | 9 | 4 | 5 | 6 | 7 | 8 | 9 | 5 | 6 | 0 | 8 | 4 | 7 | 8 | 3 |
| 1 | 4 | 0 | 1 | 2 | 3 | 4 | 5 | 1 | 1 | 1 | 3 | 4 | 5 | 0 | 1 | 2 | 3 | 4 | 5 | 0 | 1 | 2 | 3 | 4 | 5 | 1 | 2 | 4 | 4 | 0 | 3 | 4 | 1 |
| 2 | 5 | 1 | 0 | 1 | 2 | 3 | 4 | 2 | 0 | 0 | 2 | 3 | 4 | 1 | 0 | 1 | 2 | 3 | 4 | 1 | 0 | 1 | 2 | 3 | 4 | 0 | 1 | 5 | 3 | 1 | 2 | 3 | 2 |
| 3 | 6 | 2 | 1 | 0 | 1 | 2 | 3 | 3 | 1 | 1 | 1 | 2 | 3 | 2 | 1 | 0 | 1 | 2 | 3 | 2 | 1 | 0 | 1 | 2 | 3 | 1 | 0 | 6 | 2 | 2 | 1 | 2 | 3 |
| 4 | 7 | 3 | 2 | 1 | 0 | 1 | 2 | 4 | 2 | 2 | 0 | 1 | 2 | 3 | 2 | 1 | 0 | 1 | 2 | 3 | 2 | 1 | 0 | 1 | 2 | 2 | 1 | 7 | 1 | 3 | 0 | 1 | 4 |
| 5 | 8 | 4 | 3 | 2 | 1 | 0 | 1 | 5 | 3 | 3 | 1 | 0 | 1 | 4 | 3 | 2 | 1 | 0 | 1 | 4 | 3 | 2 | 1 | 0 | 1 | 3 | 2 | 8 | 0 | 4 | 1 | 0 | 5 |
| 6 | 9 | 5 | 4 | 3 | 2 | 1 | 0 | 6 | 4 | 4 | 2 | 1 | 0 | 5 | 4 | 3 | 2 | 1 | 0 | 5 | 4 | 3 | 2 | 1 | 0 | 4 | 3 | 9 | 1 | 5 | 2 | 1 | 6 |
| 7 | 3 | 1 | 2 | 3 | 4 | 5 | 6 | 0 | 2 | 2 | 4 | 5 | 6 | 1 | 2 | 3 | 4 | 5 | 6 | 1 | 2 | 3 | 4 | 5 | 6 | 2 | 3 | 3 | 5 | 1 | 4 | 5 | 0 |
| 8 | 5 | 1 | 0 | 1 | 2 | 3 | 4 | 2 | 0 | 0 | 2 | 3 | 4 | 1 | 0 | 1 | 2 | 3 | 4 | 1 | 0 | 1 | 2 | 3 | 4 | 0 | 1 | 5 | 3 | 1 | 2 | 3 | 2 |
| 9 | 5 | 1 | 0 | 1 | 2 | 3 | 4 | 2 | 0 | 0 | 2 | 3 | 4 | 1 | 0 | 1 | 2 | 3 | 4 | 1 | 0 | 1 | 2 | 3 | 4 | 0 | 1 | 5 | 3 | 1 | 2 | 3 | 2 |
| 10 | 7 | 3 | 2 | 1 | 0 | 1 | 2 | 4 | 2 | 2 | 0 | 1 | 7 | 3 | 2 | 1 | 0 | 1 | 2 | 3 | 2 | 1 | 0 | 1 | 2 | 2 | 1 | 7 | 1 | 3 | 0 | 1 | 4 |
| 11 | 8 | 4 | 3 | 2 | 1 | 0 | 1 | 5 | 3 | 3 | 1 | 0 | 1 | 4 | 3 | 2 | 1 | 0 | 1 | 4 | 3 | 2 | 1 | 0 | 1 | 3 | 2 | 8 | 0 | 8 | 1 | 0 | 5 |
| 12 | 9 | 5 | 4 | 3 | 2 | 1 | 0 | 6 | 4 | 4 | 7 | 1 | 0 | 5 | 4 | 3 | 2 | 1 | 0 | 5 | 4 | 3 | 2 | 1 | 0 | 4 | 3 | 9 | 1 | 5 | 2 | 1 | 6 |
| 13 | 4 | 0 | 1 | 2 | 3 | 4 | 5 | 1 | 1 | 1 | 3 | 4 | 5 | 0 | 1 | 2 | 3 | 4 | 5 | 0 | 1 | 2 | 3 | 4 | 5 | 1 | 2 | 4 | 4 | 0 | 3 | 4 | 1 |
| 14 | 5 | 1 | 0 | 1 | 2 | 3 | 4 | 2 | 0 | 0 | 2 | 3 | 4 | 1 | 0 | 1 | 2 | 3 | 4 | 1 | 0 | 1 | 2 | 3 | 4 | 0 | 1 | 5 | 3 | 1 | 2 | 3 | 2 |
| 15 | 6 | 2 | 1 | 0 | 1 | 2 | 3 | 3 | 1 | 1 | 1 | 2 | 3 | 2 | 1 | 0 | 1 | 2 | 3 | 2 | 1 | 0 | 1 | 2 | 3 | 1 | 0 | 6 | 2 | 2 | 1 | 2 | 3 |
| 16 | 7 | 3 | 2 | 1 | 0 | 1 | 2 | 4 | 2 | 2 | 0 | 1 | 2 | 3 | 2 | 1 | 0 | 1 | 2 | 3 | 2 | 1 | 0 | 1 | 2 | 2 | 1 | 7 | 1 | 3 | 0 | 1 | 4 |
| 17 | 8 | 4 | 3 | 2 | 1 | 0 | 1 | 5 | 3 | 3 | 1 | 0 | 1 | 4 | 3 | 2 | 1 | 0 | 1 | 4 | 3 | 2 | 1 | 0 | 1 | 3 | 2 | 8 | 0 | 8 | 1 | 0 | 5 |
| 18 | 9 | 5 | 4 | 3 | 2 | 1 | 0 | 6 | 4 | 4 | 7 | 1 | 0 | 5 | 4 | 3 | 2 | 1 | 0 | 5 | 4 | 3 | 2 | 1 | 0 | 4 | 3 | 9 | 1 | 5 | 2 | 1 | 6 |
| 19 | 4 | 0 | 1 | 2 | 3 | 4 | 5 | 1 | 1 | 1 | 3 | 4 | 5 | 0 | 1 | 2 | 3 | 4 | 5 | 0 | 1 | 2 | 3 | 4 | 5 | 1 | 2 | 4 | 4 | 0 | 3 | 4 | 1 |
| 20 | 5 | 1 | 0 | 1 | 2 | 3 | 4 | 2 | 0 | 0 | 2 | 3 | 4 | 1 | 0 | 1 | 2 | 3 | 4 | 1 | 0 | 1 | 2 | 3 | 4 | 0 | 1 | 5 | 3 | 1 | 2 | 3 | 2 |
| 21 | 6 | 2 | 1 | 0 | 1 | 2 | 3 | 3 | 1 | 1 | 1 | 2 | 3 | 2 | 1 | 0 | 1 | 2 | 3 | 2 | 1 | 0 | 1 | 2 | 3 | 1 | 0 | 6 | 2 | 2 | 1 | 2 | 3 |
| 22 | 7 | 3 | 2 | 1 | 0 | 1 | 2 | 4 | 2 | 2 | 0 | 1 | 2 | 3 | 2 | 1 | 0 | 1 | 2 | 3 | 2 | 1 | 0 | 1 | 2 | 2 | 1 | 7 | 1 | 3 | 0 | 1 | 4 |
| 23 | 8 | 4 | 3 | 2 | 1 | 0 | 1 | 5 | 3 | 3 | 1 | 0 | 1 | 4 | 3 | 2 | 1 | 0 | 1 | 4 | 3 | 2 | 1 | 0 | 1 | 3 | 2 | 8 | 0 | 8 | 1 | 0 | 5 |
| 24 | 9 | 5 | 4 | 3 | 2 | 1 | 0 | 6 | 4 | 4 | 7 | 1 | 0 | 5 | 4 | 3 | 2 | 1 | 0 | 5 | 4 | 3 | 2 | 1 | 0 | 4 | 3 | 9 | 1 | 5 | 2 | 1 | 6 |
| 25 | 5 | 1 | 0 | 1 | 2 | 3 | 4 | 2 | 0 | 0 | 2 | 3 | 4 | 1 | 0 | 1 | 2 | 3 | 4 | 1 | 0 | 1 | 2 | 3 | 4 | 0 | 1 | 5 | 3 | 1 | 2 | 3 | 2 |
| 26 | 6 | 2 | 1 | 0 | 1 | 2 | 3 | 3 | 1 | 1 | 1 | 2 | 3 | 2 | 1 | 0 | 1 | 2 | 3 | 2 | 1 | 0 | 1 | 2 | 3 | 1 | 0 | 6 | 2 | 2 | 1 | 2 | 3 |
| 27 | 0 | 4 | 5 | 6 | 7 | 8 | 9 | 3 | 5 | 5 | 7 | 8 | 9 | 4 | 5 | 6 | 7 | 8 | 9 | 4 | 5 | 6 | 7 | 8 | 9 | 5 | 6 | 0 | 8 | 4 | 7 | 8 | 3 |
| 28 | 8 | 4 | 3 | 2 | 1 | 0 | 1 | 5 | 3 | 3 | 1 | 0 | 1 | 4 | 3 | 2 | 1 | 0 | 1 | 4 | 3 | 2 | 1 | 0 | 1 | 3 | 2 | 8 | 0 | 4 | 1 | 0 | 5 |
| 29 | 4 | 0 | 1 | 2 | 3 | 4 | 5 | 1 | 1 | 1 | 3 | 8 | 5 | 0 | 1 | 2 | 3 | 8 | 5 | 0 | 1 | 2 | 3 | 8 | 5 | 1 | 2 | 4 | 4 | 0 | 3 | 4 | 1 |
| 30 | 7 | 3 | 2 | 1 | 0 | 1 | 2 | 4 | 2 | 2 | 0 | 1 | 2 | 3 | 2 | 1 | 0 | 1 | 2 | 3 | 2 | 1 | 0 | 1 | 2 | 2 | 1 | 7 | 1 | 3 | 0 | 7 | 4 |
| 31 | 8 | 4 | 3 | 2 | 1 | 0 | 1 | 5 | 3 | 3 | 1 | 0 | 1 | 4 | 3 | 2 | 1 | 0 | 1 | 4 | 3 | 2 | 1 | 0 | 1 | 3 | 2 | 8 | 0 | 4 | 7 | 0 | 5 |
| 32 | 3 | 1 | 2 | 3 | 4 | 5 | 6 | 0 | 2 | 2 | 4 | 5 | 6 | 1 | 2 | 3 | 4 | 5 | 6 | 1 | 2 | 3 | 4 | 5 | 6 | 2 | 3 | 3 | 5 | 1 | 4 | 5 | 0 |